\documentclass{article}

\usepackage{microtype}
\usepackage{graphicx}
\usepackage{subfigure}
\usepackage{booktabs} %

\usepackage{hyperref}

\usepackage[ruled]{algorithm2e} %

\usepackage[accepted]{icml2024}

\usepackage{amsmath}
\usepackage{amssymb}
\usepackage{mathtools}
\usepackage{amsthm}

\usepackage{ulem}
\usepackage{multirow}
\usepackage{multicol}
\usepackage{makecell}
\usepackage{adjustbox}
\usepackage{bm}
\usepackage{xcolor} 
\usepackage{colortbl}

\usepackage[justification=justified,skip=5pt]{caption}
\usepackage{pifont}

\usepackage[capitalize,noabbrev]{cleveref}

\definecolor{softred}{RGB}{239,154,154}    %
\definecolor{softblue}{RGB}{173,216,230}   %
\definecolor{softgreen}{RGB}{162,205,90}   %
\definecolor{softyellow}{RGB}{255,255,224} %
\definecolor{darkred}{RGB}{139,0,0}       %
\definecolor{darkblue}{RGB}{0,0,139}      %
\definecolor{darkgreen}{RGB}{0,100,0}     %
\definecolor{darkorange}{RGB}{255,140,0}  %

\makeatletter
\g@addto@macro\normalsize{%
  \abovedisplayskip 4pt plus 2pt minus 3pt%
  \belowdisplayskip \abovedisplayskip
  \abovedisplayshortskip 4pt plus2pt  minus3pt%
  \belowdisplayshortskip 4pt plus2pt minus3pt%
}

\makeatother

\theoremstyle{plain}

\theoremstyle{definition}

\theoremstyle{remark}

\usepackage[textsize=tiny]{todonotes}

\icmltitlerunning{MITA: Bridging the Gap between Model and Data for Test-time Adaptation}

\begin{document}

\twocolumn[
\icmltitle{MITA: Bridging the Gap between Model and Data for Test-time Adaptation}

\icmlsetsymbol{equal}{*}

\begin{icmlauthorlist}
\icmlauthor{Yige Yuan}{ict,ucas}
\icmlauthor{Bingbing Xu}{ict}
\icmlauthor{Teng Xiao}{psu}
\icmlauthor{Liang Hou}{ks}
\icmlauthor{Fei Sun}{ict}
\icmlauthor{Huawei Shen}{ict,ucas}
\icmlauthor{Xueqi Cheng}{ict,ucas}
\\
\begin{small}
$^{1}$CAS Key Laboratory of AI Safety, Institute of Computing Technology, Chinese Academy of Sciences\\
$^{2}$University of Chinese Academy of Sciences 
$^{3}$Pennsylvania State University
$^{4}$Kuaishou Technology\\
\{yuanyige20z, xubingbing, sunfei, shenhuawei, cxq\}@ict.ac.cn, tengxiao@psu.edu, lianghou96@gmail.com
\vspace{-4mm}
\end{small}
\end{icmlauthorlist}

\icmlkeywords{Machine Learning, ICML}

\vskip 0.3in
]

\begin{abstract}

Test-Time Adaptation (TTA) has emerged as a promising paradigm for enhancing the generalizability of models. 
However, existing mainstream TTA methods, predominantly operating at batch level, often exhibit suboptimal performance in complex real-world scenarios, particularly when confronting outliers or mixed distributions.
This phenomenon stems from a pronounced over-reliance on statistical patterns over the distinct characteristics of individual instances, resulting in a divergence between the distribution captured by the model and data characteristics.
To address this challenge, we propose Meet-In-The-Middle based Test-Time Adaptation (MITA), which introduces energy-based optimization to encourage mutual adaptation of the model and data from opposing directions, thereby meeting in the middle.
MITA pioneers a significant departure from traditional approaches that focus solely on aligning the model to the data, facilitating a more effective bridging of the gap between model's distribution and data characteristics.
Comprehensive experiments with MITA across three distinct scenarios (Outlier, Mixture, and Pure) demonstrate its superior performance over SOTA methods, highlighting its potential to significantly enhance generalizability in practical applications.

\end{abstract}

\section{Introduction}

\begin{figure}[!t]
    \centering
    \includegraphics[width=0.85\linewidth]{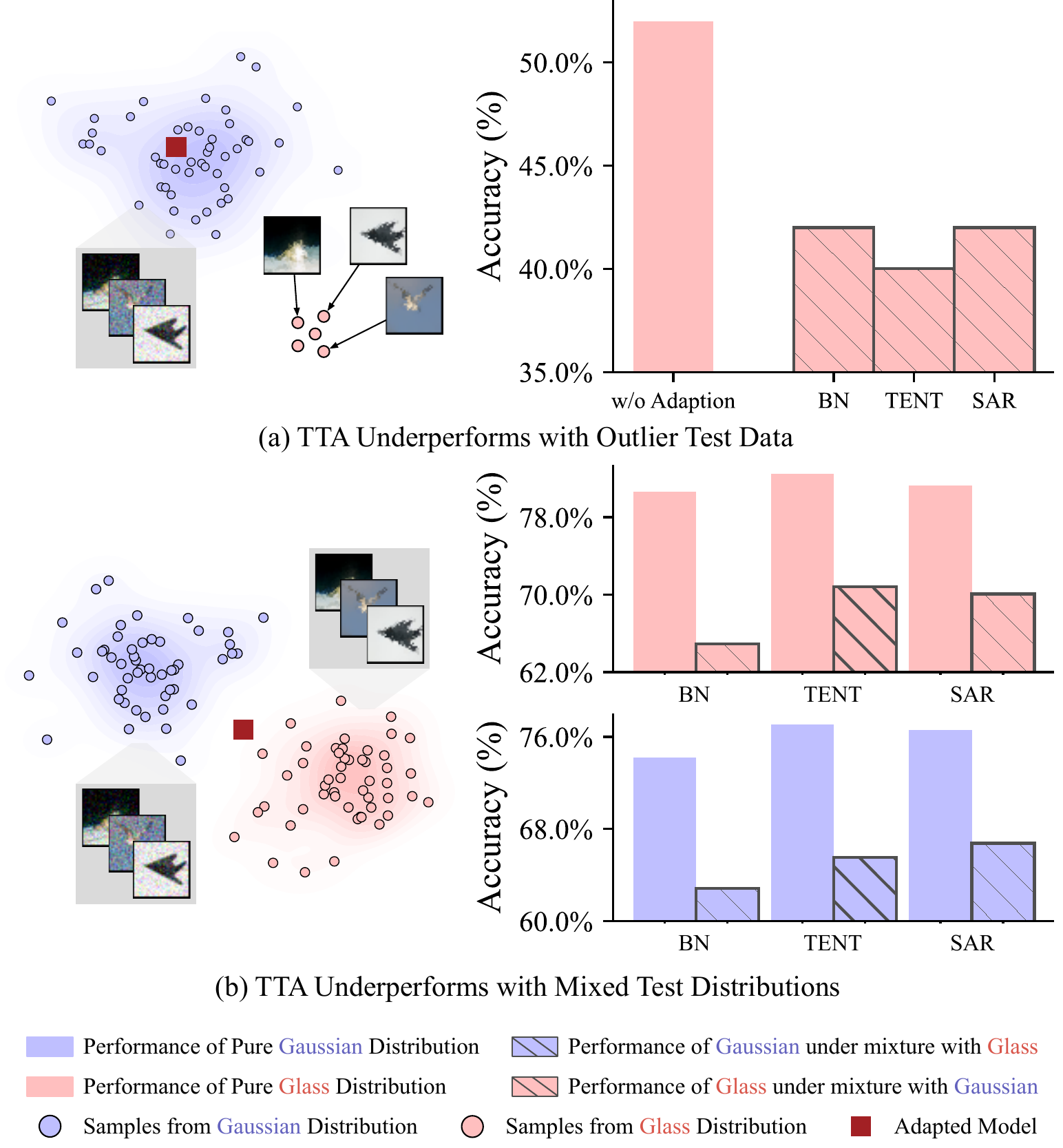}
    \caption{Batch-level TTA performance noticeably declines in the presence of outliers or mixed distributions.}
    \label{fig:motivation}
    \vspace{-3mm}
\end{figure}

Test-Time Adaptation (TTA)~\cite{liang2023comprehensive} increasingly emerges as a promising paradigm, offering an effective solution to enhance generalization under distribution shifts~\cite{jordan2015machine}.
TTA uses unlabeled test data to enhance the generalizability of a trained model during the test phase, eliminating the need for accessing training data and processes, which is particularly beneficial for large, open-source models where training details are often proprietary or resource-constrained~\cite{touvron2023llama}.

\begin{figure*}[!t]
    \centering
    \includegraphics[width=\linewidth]{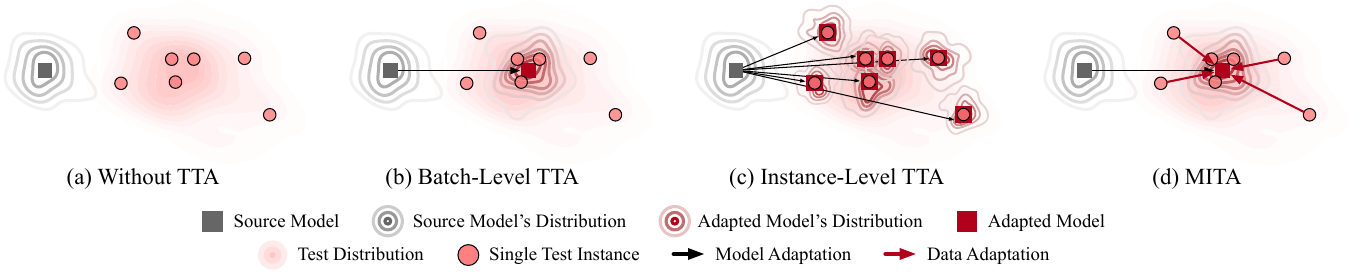}
    \vspace{-4mm}
    \caption{Four TTA paradigms: The model have better generalizability for data that aligns with the model's distribution. MITA encourages mutual adaptation of the model and data from opposing directions, thereby meeting in the middle.
    }
    \label{fig:difference}
\end{figure*}

Existing TTA methods typically fall into two categories: batch-level adaptation and instance-level adaptation. 
The prevalent paradigm, batch-level adaptation, tunes a trained model to align with the statistical patterns of a batch of test samples, guided by objectives such as aligning the statistics in BatchNorm~\cite{ioffe2015batch,schneider2020improving}.
However, they often exhibit suboptimal performance in complex real-world scenarios, particularly when confronting outliers or mixed distributions.
As illustrated in Figure~\ref{fig:motivation}(a), in the presence of outlier samples within a predominant distribution, the performance of all investigated methods declines, falling below that achieved by the unadapted source model specifically for these outlier instances.
Similarly phenomenon is shown in Figure~\ref{fig:motivation}(b), where two distributions are uniformly mixed.
This issue arises due to the model's over-reliance on overall statistical patterns rather than the unique attributes of each instance, leading to a discrepancy between the distribution captured by the model and the actual distribution of test data.

Instance-level adaptation~\cite{zhang2022memo} can mitigate the above limitations, which target the unique characteristics of instances by tuning the model for each instance respectively. 
However, it is computationally demanding~\cite{niu2022efficient} and may exhibit suboptimal performance due to its inability to access statistical knowledge~\cite{niu2023towards}.
Overall, \textit{how to bridge the gap between model and data}, i.e., efficiently aligning the model with instance-specific characteristics while still leveraging the advantages of statistical knowledge, remains an open question.

To tackle the above challenge, we propose \textbf{M}eet-\textbf{I}n-The-Middle based \textbf{T}est Time \textbf{A}daptation, namely \textbf{MITA}, 
a method that enables the model and data to undergo mutual adaptation from opposite directions, thereby encouraging them to meet in the middle.
Specifically, MITA reinterprets the source model as an energy-based model and then conducts the following two parts: model adaptation and data adaptation.
Model adaptation uses Contrastive Divergence~\cite{hinton2002training} as the adaptation objective to infuse the model with a perception of the test data distribution. 
The gained perception enables the model to be equipped with generative capabilities.
For data adaptation, leveraging the newfound generative ability, we introduce a dynamic self-update mechanism for each instance within the batch via Langevin Dynamics~\cite{welling2011bayesian}, making it further align with the model.
Based on these, MITA can not only maintain the statistical knowledge and efficiency inherent in batch-level TTA but also perceive each instance.

As illustrated in~\cref{fig:difference}, in contrast to previous batch-level and instance-level TTA, which focuses solely on aligning the model with the data, MITA pioneers an innovative paradigm of mutual adaptation to bridge the gap between model and data.
Extensive experiments conducted with MITA in three distinct scenarios (Outlier, Mixture, and Pure) show its superior performance compared to SOTA methods. The comprehensive ablation studies and visualization further highlight its potential to significantly enhance generalizability.

Our main contributions include:

\begin{itemize}
\vspace{-3mm}
\item \textbf{A promising paradigm}: To the best of our knowledge, we are the first to pioneer mutual adaptation paradigm, a significant departure from traditional approaches that focus solely on aligning the model to the data.
\item \textbf{An innovative method}: We propose MITA which introduces energy-based optimization to encourage mutual adaptation of the model and data from opposing directions, thereby meeting in the middle.
\item \textbf{Solid experiments}: Extensive experiments reveal that MITA outperforms baselines across three distinct scenarios, e.g., the improvements are up to 10.57\% in Outlier and 4.68\% in Mixter.
\end{itemize}

\section{Preliminary}

\subsection{Test-Time Adaptation}

Let $\{(\mathbf{x}_{\text{train}}, y_{\text{train}})\} \subset \mathcal{X} \times \mathcal{Y}$ be a set of labeled training data with a distribution of $p_{\mathrm{train}}(\mathbf{x},y)$, where $\mathcal{X}$ and $\mathcal{Y}$ are data and label spaces. 
Let $f_{\theta}: \mathcal{X} \rightarrow \mathcal{Y}$ be a source model parameterized by $\theta$ that is trained on the training dataset. 
The source model, optimized to fit the training dataset, is designed to perform well on the test data $\mathbf{x}_\mathrm{test}\sim p_{\text{test}}(\mathbf{x})$ that follows the same distribution, i.e., $p_{\text{test}}(\mathbf{x}) \approx p_{\text{train}}(\mathbf{x})$. 
Nonetheless, its performance can be poor and unreliable when dealing with test data that does not follow the training distribution, i.e., $p_{\text{test}}(\mathbf{x}) \neq p_{\text{train}}(\mathbf{x})$.

Test-time adaptation aims to enhance the source model's generalizability on specific test data through unsupervised fine-tuning. 
Specifically, given the adapting objective $\mathcal{L}$, TTA of model $f_{\theta}$ on test data $\mathbf{x}_{\mathrm{test}}$ can be formulated as,
\begin{equation}
\min_{\theta} \mathbb{E}_{\mathbf{x}_{\mathrm{test}}} \mathcal{L}(\mathbf{x}_{\mathrm{test}} ; \theta)
\label{eq:tta}
\end{equation}

\subsection{Energy-based Model}
\label{sec:preliminary_ebm}

Energy-Based Model (EBM)~\cite{lecun2006tutorial,song2021train} is a type of probabilistic model characterized by an energy function $E_{\theta}: \mathcal{X} \rightarrow \mathbb{R}$ that outputs an energy value for any given data $\mathbf{x} \in \mathcal{X}$. 
To express the density $p(\mathbf{x})$, EBM utilizes the the Boltzmann distribution~\cite{lifshitz1980statistical} defined by
\begin{equation}
    p_{\theta}(\mathbf{x})=\frac{\exp \left(-E_{\theta}(\mathbf{x})\right)}{Z_{\theta}}.
    \label{eq:ebm}
\end{equation}
The objective of EBM is to learn the distribution of $\mathbf{x}$ by optimizing $\theta$. 
The energy value of each sample can be viewed as an unnormalized probability, with lower scores denoting higher likelihoods~\cite{du2019implicit}.
The partition function $Z_{\theta} = \int \exp(-E_{\theta}(\mathbf{x})) \, \mathrm{d}\mathbf{x}$ serves to normalize the probability density ($\int p_{\theta}(\mathbf{x}) \mathrm{d}\mathbf{x}=1$).

Directly optimizing~\cref{eq:ebm} is computationally intractable as the partition function $ Z_{\theta}$ necessitates integration over the whole high-dimensional data space of $\mathbf{x}$.
To overcome this difficulty, Contrastive Divergence~\cite{hinton2002training} had been proposed to estimate the gradient of the log-likelihood,
\begin{equation}
\frac{\partial \log p_\theta(\mathbf{x})}{\partial \theta}=\mathbb{E}_{\tilde{\mathbf{x}} \sim p_\theta}\left[\frac{\partial E_\theta\left(\tilde{\mathbf{x}}\right)}{\partial \theta}\right]-\frac{\partial E_\theta(\mathbf{x})}{\partial \theta}.
\label{eq:cd}
\end{equation}
Assuming the model is optimized perfectly, one can sample new data through Stochastic Gradient Langevin Dynamics (SGLD)~\cite{welling2011bayesian}.
As is shown in \cref{eq:sgld}, starting from an initial noise distribution $p_0$, SGLD runs $T$ random gradient descent to transfer the data to follows the model distribution $p_\theta$,
\begin{equation}
\small
\tilde{\mathbf{x}}_{t+1}=\mathbf{x}_t-\frac{\alpha}{2} \frac{\partial E_\theta\left(\tilde{\mathbf{x}}_t\right)}{\partial \tilde{\mathbf{x}}_t}+\sqrt{\alpha}\boldsymbol{\epsilon}, \, \boldsymbol{\epsilon} \sim \mathcal{N}(\boldsymbol{0}, \mathbf{I}) , \, \tilde{\mathbf{x}}_{0} \sim p_0,
\label{eq:sgld}
\end{equation}
where $\alpha\in\mathbb{R}$ denotes the step-size.

Unlike most other probabilistic models, 
EBMs do not necessitate the normalizing constant to be tractable and do not require an explicit neural network for distribution modeling and sample generation, implying the generation process is implicit~\cite{du2019implicit}.
These lead to increased flexibility in parameterization and allow for modeling a wider range of probability distributions.

\section{Method}
\begin{figure*}[!t]
    \centering
    \includegraphics[width=\linewidth]{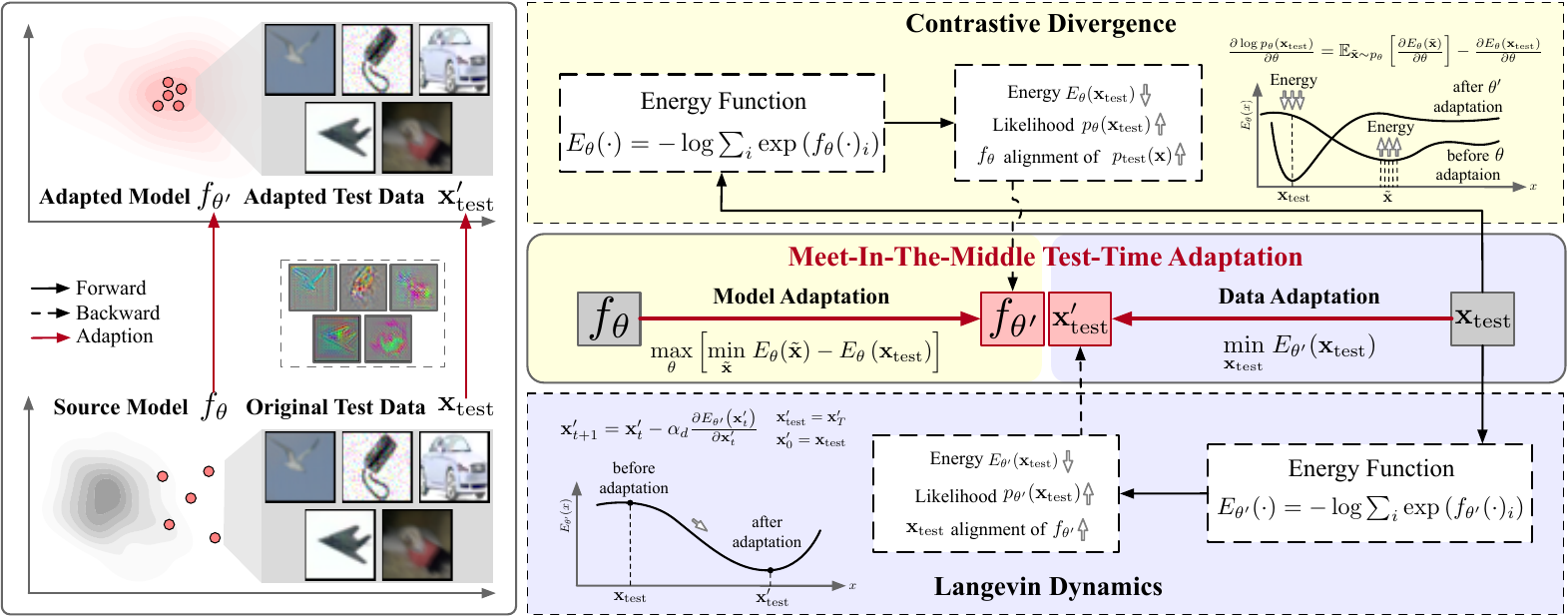}
    \vspace{-3mm}
    \caption{
\textbf{Overview of MITA}.
Left: the motivation of MITA.
Right: the overall architecture of MITA, which establishes a mutual adaptation between a trained source model and the test data, guiding both to meet in the middle.}
    \label{fig:model}
\end{figure*}

In this section, we detail our method, Meet-In-The-Middle Test-Time Adaptation (MITA).
We begin with the motivation and overall framework in~\cref{sec:overall_arch}, 
followed by a breakdown of the method's components in~\cref{sec:energy_based_model_construction,sec:model_ada,sec:data_ada}. 
Finally, a comparative discussion of MITA and its related paradigms is provided in~\cref{sec:method_disc}.

\subsection{Motivation and Overall Framework}
\label{sec:overall_arch}

The overview of MITA is depicted in~\cref{fig:model}. 
The motivation is illustrated in the left side: 
Initially, the model and data are completely unaligned, i.e., the data distribution modeled captured by the model is inconsistent with the test data, resulting in poor generalization performance.
With adaptation, the model and data achieve \textbf{mutual alignment}, where the model aligns to the data distribution as a whole, and each data instance aligns with the adapted model in turn.
To achieve this, we reinterpret the source-trained model as an \textbf{energy-based model}. 
Building upon this, we separately optimize the model through \textbf{model adaptation} and the data through \textbf{data adaptation} to maximize the likelihood of the target test data under the model's distribution, as illustrated in the right side.  
In the following, we provide a detailed introduction to the construction of the energy-based model and two adaptation components.

\subsection{Energy-Based Model Construction}
\label{sec:energy_based_model_construction}

The reason we treat the source-trained model as an EBM is that EBM possesses two capabilities: 1) adjusting the captured distribution behind one model and 2) generating samples to satisfy the given distribution, as introduced in \ref{sec:preliminary_ebm}.
Due to this flexibility, we can integrate the distribution modeling and generative capabilities of $\mathbf{x}$ into a discriminative architecture~\cite{Grathwohl2020Your,du2019implicit,han2019divergence}.
The integration exactly matches our mutual adaptation goal under TTA scenario:
(1) enabling model adaptation, i.e., making the model align with a collection of test data, through enhancing the model's perception of the distribution.
(2) enabling data adaptation, i.e., making each instance further align with the adapted model respectively, through adjusting data guided by the distribution-perception ability of the model.
Such a mutual adaptation can not only push the model’s captured distribution toward the target data but also pull the data toward the model, making the data fall within the model’s captured distribution, thus improving the generalization ability.

To treat the source model $f_{\theta}$ as an EBM, we reinterpret its logits to construct one.
This construction is based on the fundamental understanding that an energy-based framework is inherent to any discriminative model~\cite{Grathwohl2020Your}.
In this framework, the energy of one sample for a corresponding class can be represented as its logit produced by the trained classifier, denoted by $E_{\theta}(\mathbf{x},y) = -f_\theta(\mathbf{x})[y]$.
Therefore, following the definition in \cref{eq:ebm}, the joint probability distribution of $\mathbf{x}$ and $y$ can be defined as,
\begin{equation}
p_\theta(\mathbf{x}, y) = \frac{\exp(f_\theta(\mathbf{x})[y])}{Z_\theta},
\label{eq:p1}
\end{equation}
where $Z_\theta=\int\sum_y\exp(f_\theta(\mathbf{x})[y])\mathrm{d}\mathbf{x}$.
Then the marginal distribution of $\mathbf{x}$ can be obtained by marginalizing over $y$, as shown below,
\begin{equation}
p_\theta(\mathbf{x}) = \sum_y p_\theta(\mathbf{x}, y) = \frac{\sum_y \exp(f_\theta(\mathbf{x})[y])}{Z_\theta}.
\label{eq:p2}
\end{equation}
By substituting \cref{eq:p2} into \cref{eq:ebm}, we can obtain the form of the energy function as follows:
\begin{equation}
E_\theta(\mathbf{x})=-\log \sum_y \exp \left(f_\theta(\mathbf{x})[y]\right)
\label{eq:lse_energy}.
\end{equation}
Following the aforementioned steps, we repurpose and reinterpret the logits produced by the trained classifier to establish an energy-based model and define the energy function as the negative log-sum-exp of the logits.

\subsection{Model Adaptation}
\label{sec:model_ada}

As introduced in \cref{sec:energy_based_model_construction}, constructing the energy-based model provides the foundation of model adaptation, which aims to enable a model to capture a target distribution, thereby providing a solution to a fundamental challenge in test-time adaptation: covariate shift~\cite{jiang2008literature,yuan2023tea}. 
Our MITA performs model adaptation by minimizing the energy of test data, i.e., maximizing the likelihood of test data in the EBM via Contrastive Divergence, as shown in \cref{eq:cd}.
This objective can be fundamentally understood as a min-max game as shown in \cref{eq:minmax}, which minimizes the energy derived from the incoming test samples while concurrently amplifying the energy of fictitious samples obtained via SGLD from the classifier's distribution.

\begin{equation}
    \theta^{\prime} = \arg \underset{\theta}{\max} \mathbb{E}_{\mathbf{x}} \left[ \underset{\tilde{\mathbf{x}}}{\min} \; E_{\theta}(\tilde{\mathbf{x}}) - E_{\theta}\left(\mathbf{x}_{\mathrm{test}}\right) \right]
    \label{eq:minmax}
\end{equation}

By adapting through this objective, the trained source classifier gradually aligns with the test data distribution, thereby bolstering the model's perception of the test distribution and thus enhancing generalizability.
The pseudocode for model adaptation can be found in~\cref{alg:model}.

However, despite the enhanced generalization performance, this adaptation still heavily relies on the overall patterns of the distribution from a collection of samples, which may overlook the characteristics of individual instances, resulting in the limited generalization (see \cref{fig:motivation}).

\begin{algorithm}[!t]
\setcounter{AlgoLine}{0}
\caption{Model Adaptation ($\mathrm{ModelAda}$)}
  \label{alg:model}
  \KwIn{Classifier $f_{\theta}$; Test Samples $\mathbf{x}_{\mathrm{test}}$; Langevin Sampling Step Size $\alpha$; Langevin Sampling Steps $T$; Noise Distribution $p_0$; Model Adaptation Rate $\beta$; Model Adaptation Steps $N$.}
  \KwOut{Parameters after adaptation $\theta$}
  $E_\theta(\cdot) \leftarrow -\log \sum_y \exp \left(f_\theta(\cdot)[y]\right)$
  
  \For{$i \leftarrow  0,1,\dots,N-1$}{

    $\tilde{\mathbf{x}}_0 \leftarrow  \mathrm{sample}(p_0)$
    
    \For{$t \leftarrow  0,1,\dots,T-1$}{

        $\boldsymbol{\epsilon} \leftarrow \mathrm{sample}(\mathcal{N}(\boldsymbol{0}, \mathbf{I}))$
        
        $\tilde{\mathbf{x}}_{t+1}\leftarrow \tilde{\mathbf{x}}_t-\frac{\alpha}{2} \frac{\partial E_\theta\left(\tilde{\mathbf{x}}_t\right)}{\partial \tilde{\mathbf{x}}_t}+\sqrt{\alpha}\boldsymbol{\epsilon}$

    }

    $\tilde{\mathbf{x}} \leftarrow  \tilde{\mathbf{x}}_{T-1}$

    $ \theta \leftarrow \theta -\beta  \nabla_{\theta} \left[ E_{\theta}(\mathbf{x}_{\mathrm{test}}) - E_{\theta}(\tilde{\mathbf{x}}) \right]$
  }
 \Return{$\theta$}
 
\end{algorithm}

\subsection{Data Adaptation}
\label{sec:data_ada}

\begin{algorithm}[!t]
\setcounter{AlgoLine}{0}
\caption{Data Adaptation ($\mathrm{DataAda}$)}
  \label{alg:data}
  \KwIn{Classifier $f_{\theta}$; Test Samples $\mathbf{x}_{\mathrm{test}}$; Data Adaptation Step Size $\alpha_d$; Data Adaption Step $T_d$.}
  \KwOut{Data after adaptation $\mathbf{x}_{\mathrm{test}}$}
    
    $E_\theta(\cdot) \leftarrow -\log \sum_y \exp \left(f_\theta(\cdot)[y]\right)$
    
    $\tilde{\mathbf{x}}_0 \leftarrow \mathbf{x}_{\mathrm{test}}$
    
    \For{$t \leftarrow  0,1,\dots,T_d-1$}{

        $\tilde{\mathbf{x}}_{t+1}\leftarrow \tilde{\mathbf{x}}_t-\frac{\alpha_{d}}{2} \frac{\partial E_\theta\left(\tilde{\mathbf{x}}_t\right)}{\partial \tilde{\mathbf{x}}_t}$ %

    }

    $\mathbf{x}_{\mathrm{test}} \leftarrow  \tilde{\mathbf{x}}_{T-1}$
    
    \Return{$\mathbf{x}_{\mathrm{test}}$}
 
\end{algorithm}

It is worth noting that model adaptation incorporates the marginal distribution $p(\mathbf{x})$ into the classifier, thus improving the model's generative capability for $\mathbf{x}$.
Such capability makes this classifier different from traditional classifiers which focus on the conditional distribution $p(y|\mathbf{x})$ only, paving the way for data adaptation by actively aligning each instance with the adapted distribution of the model $p_{\theta^{\prime}}(\mathbf{x})$.

\begin{table*}[t]
\centering
\caption{Comparison of our method and relevant test-time adaptation paradigm}
\begin{adjustbox}{width=\linewidth}
\setlength{\tabcolsep}{3mm}
\begin{tabular}{l|cccccc}
\toprule
 Paradigm & Test Adaptation Object & Training Data Independency & Efficiency & Collective Knowledge & Instance Alignment\\
\midrule
Batch-Level Adaptation & Model & \ding{52} & \ding{52} & \ding{52} & \ding{56}  \\
Instance-Level Adaptation & Model & \ding{52} & \ding{56} & \ding{56} & \ding{52} \\
Purification-Based Adaptation & Data & \ding{56} & \ding{52} & \ding{56} & \ding{52} \\
\midrule
MITA (ours) & Model \& Data & \ding{52} & \ding{52} & \ding{52} & \ding{52}  \\
\bottomrule
\end{tabular}
\end{adjustbox}
\label{tab:summary}
\end{table*}

The data adaptation is also rooted in energy-based optimization, as introduced in \cref{sec:energy_based_model_construction}.
However, unlike model adaptation, which constructs an energy-based model $E_{\theta}$ from the source model $f_{\theta}$ and optimizes the model parameters to minimize the average energy over all data, data adaptation builds an energy-based model $E_{\theta^{\prime}}$ upon the adapted model $f_{\theta^{\prime}}$ and optimizes the data itself to minimize energy of each data, i.e., maximize likelihood, as is shown below.
\begin{equation}
\begin{split}
E_{\theta^{\prime}}(\mathbf{x}) &=-\log \sum_y \exp \left(f_{\theta^{\prime}}(\mathbf{x})[y]\right) \\
\mathbf{x}_{\mathrm{test}}^{\prime} &= \arg \underset{\mathbf{x}_{\mathrm{test}}}{\min} \; E_{\theta^{\prime}}(\mathbf{x}_{\mathrm{test}})
\label{eq:data_obj}
\end{split}
\end{equation}
This process can be achieved via SGLD with the initialization of the chain being the original test data, $\mathbf{x}_{\mathrm{test}}$.
Starting from each original test data point, SGLD iteratively moves the point towards a region of lower energy that is more consistent with the distribution embedded by the model. 
Notably, we remove the noise term from the SGLD, as the data does not require randomness, which is demonstrated below.
\begin{equation}
\tilde{\mathbf{x}}_{t+1}  =\mathbf{x}_t-\frac{\alpha_d}{2} \frac{\partial E_{\theta^{\prime}}\left(\tilde{\mathbf{x}}_t\right)}{\partial \tilde{\mathbf{x}}_t},\quad \text{where} \quad \tilde{\mathbf{x}}_{0} = \mathbf{x}_{\mathrm{test}},
\label{eq:data_sgld}
\end{equation}
where $\alpha_d$ represents the step size and $t=1,2,\dots,T_d$ denotes the data iteration step. 
The pseudocode for data adaptation can be found in~\cref{alg:data}.

Notably, in our implementation, we do not use the adapted model $f_{\theta^{\prime}}$ to adapt data directly. Instead, we train a new model, $f_{\tilde{\theta}^{\prime}}$ in the exactly same way as $f_{\theta^{\prime}}$, with the key difference being the number of epochs for model adaptation. 
The new model $f_{\tilde{\theta}^{\prime}}$ undergoes more epochs than $f_{\theta^{\prime}}$, indicating a more thorough energy-based optimization.

The reason behind this implementation is grounded in the trade-off between a model's discriminability and its transferability, as discussed in prior research~\cite{kundu2022balancing}. 
In the context of MITA, the distribution modeling and generative abilities of the test data represent transferability.
Specifically, over-optimization with respect to the unsupervised adaptation objective can undermine the model's discriminative performance. 
Conversely, a model that has undergone only slight adaptation, while maintaining its discriminability, may not possess the necessary generative capabilities for data adaptation. This phenomenon has also been confirmed in \cref{tab:ablation}.
This implementation allows $f_{\theta^{\prime}}$ to emphasize discriminative tasks, facilitating inference on test data for the target task, while $f_{\tilde{\theta}^{\prime}}$ focuses on data adaptation to align it with the distribution embedded in the model.
The pseudocode for MITA can be found in~\cref{alg:mita}.

\begin{algorithm}[h]
\setcounter{AlgoLine}{0}
\caption{Meet-In-The-Middle Test-time Adaptation}
  \label{alg:mita}
  \KwIn{Trained Classifier $f_{\theta}$; Test Samples $\mathbf{x}_{\mathrm{test}}$.}
  \KwOut{Predictions for all $\mathbf{x}_{\mathrm{test}}$}

  $\theta^{\prime}=\mathrm{ModelAda}(f_{\theta}, \mathbf{x}_{\mathrm{test}})$
  
  $\tilde{{\theta}}^{\prime}=\mathrm{ModelAda}(f_{\theta}, \mathbf{x}_{\mathrm{test}})$

  $\mathbf{x}_{\mathrm{test}}^{\prime}=\mathrm{DataAda}(f_{\tilde{{\theta}}^{\prime}}, \mathbf{x}_{\mathrm{test}})$
  
 \Return{$f_{\theta^{\prime}}(\mathbf{x}_{\mathrm{test}}^{\prime})$}
 
\end{algorithm}

\subsection{Discussion}
\label{sec:method_disc}

In this section, we delve into the relationship and difference between our MITA and relevant test-time adaptation paradigms, with a comparison summarized in \cref{tab:summary}.

\paragraph{Batch-Level Adaptation}
Batch-Level Adaptation is the mainstream TTA approach. 
Representative methods include 
BN~\cite{schneider2020improving}, adapting BatchNorm statistics with test data; 
TENT~\cite{wang2021tent}, fine-tuning BatchNorm layers via test phase entropy minimization; 
EATA~\cite{niu2022efficient}, employing a Fisher regularizer to prevent excessive parameter changes; 
SAR~\cite{niu2023towards}, removing high-gradient samples and promoting flat-minima weights; 
and SHOT~\cite{liang2020we}, combining entropy minimization with pseudo-labeling.
Recently, a method called TEA~\cite{yuan2023tea} has been proposed, which enhances model generalizability via energy-based optimization. 
TEA focuses solely on energy-based model adaptation, which can be viewed as half of MITA framework, i.e., without the component of data adaptation.
However, these methods still depend on the overall patterns of a collection of samples and overlook instance-level alignment.

\paragraph{Instance-Level Adaptation}
Instance-Level Adaptation refers to a TTA paradigm that can conduct adaptation for a single instance without needing other samples. Representative methods include MEMO~\cite{zhang2022memo} and TTT~\cite{pmlr-v119-sun20b}, which generate a batch of augmented data from a single sample using various data augmentation techniques, and then adapt the model for each sample based on its augmentations. 
These methods, while recognizing individual instance attributes, incur high computational costs by requiring model retraining for each test instance and fail to use shared knowledge from batched or previous samples, potentially leading to limited performance.

\paragraph{Purification-Based Adaptation}
Diffusion-Driven Adaptation (DDA)~\cite{Gao_2023_CVPR} is a purification-based TTA method, which proposes to update the test data. 
This method involves learning a diffusion model on the source data, and then projecting test data back to the source domain. 
DDA solely performs data adaptation to align the test data with the source data distribution unidirectionally, relying on the training data and training process, which may be unavailable for many real-world scenarios.
MITA utilizes energy-based adaptation to perform both mutual adaptation during the testing phase to endow the model with generative capabilities, avoiding the need for a well-trained diffusion model on the training data, which is more practical for wider scenarios.

\section{Experiment}

We conduct evaluations on three aspects:
(1) Comparison of MITA with state-of-the-art methods across three practical scenarios (\cref{4.2}).
(2) Ablation studies to validate the effectiveness of each component (\cref{4.3}).
(3) Visualization of the adaptation processes(\cref{4.4}).

Due to space limitations, additional scenarios and further analyses are provided in~\cref{app:exp}, including
(1) Extensive mixture ratios and distributions (\cref{app:exp_mix}).
(2) A wide range of visualization results (\cref{app:exp_vis}).
(3) Computing complexity analysis (\cref{app:exp_complexity}).

\begin{table*}[!t]
\renewcommand{\arraystretch}{1}
\caption{Comparisons of MITA and baselines on outlier test data and mixed test distributions using CIFAR-10-C. Colors indicate four different corruption categories: Red for Noise, Green for Blur, Blue for Weather, and Orange for Digital. Adaptations that degrade the source model performance are marked with $\ast$. The best  results are highlighted in \textbf{boldface}.}
\centering
\begin{adjustbox}{width=\textwidth}
\setlength{\tabcolsep}{1mm}
\begin{tabular}{clcccccccccccccccccc}
\toprule
\multicolumn{2}{c}{\multirow{2.5}{*}{\centering Dist. A / Dist. B}}
& \multicolumn{3}{c}{\textcolor{darkgreen}{Glass} / \textcolor{darkred}{Guass.} } 
& \multicolumn{3}{c}{\textcolor{darkblue}{Pixe.} / \textcolor{darkred}{Shot} } 
& \multicolumn{3}{c}{\textcolor{darkgreen}{Moti.} / \textcolor{darkblue}{Contr.}} 
& \multicolumn{3}{c}{\textcolor{darkblue}{Elast.} / \textcolor{darkorange}{Fog}} 
& \multicolumn{3}{c}{\textcolor{darkred}{Impul.}/ \textcolor{darkred}{Gauss.}} 
& \multicolumn{3}{c}{\textcolor{darkgreen}{Defoc.}/ \textcolor{darkgreen}{Zoom}} \\
\cmidrule(lr){3-5}
\cmidrule(lr){6-8}
\cmidrule(lr){9-11}
\cmidrule(lr){12-14}
\cmidrule(lr){15-17}
\cmidrule(lr){18-20}
&
& Glass & Guass.  & All
& Pixe. & Shot & All
& Moti. & Contr. & All
& Elast. & Fog & All
& Impul. & Gauss. & All
& Defoc. & Zoom  & All\\
\midrule
\multirow{9}*{\parbox[t]{0.2cm}{\centering \rotatebox{90}{Outlier}}} 
& \cellcolor{pink!20}Source & 52.00 & 27.67 & 27.79 & 50.00 & 34.26 & 34.34 & 64.00 & 53.35 & 53.41 & 76.00 & 73.96 & 73.98 & 38.00 & 27.66 & 27.72  & 50.00 & 57.95 & 57.91 \\
& BN & 42.00${\ast}$ & 71.92 & 71.77 & 34.00${\ast}$ & 73.82 & 73.62 & 48.00${\ast}$ & 87.41 & 87.22 & 62.00${\ast}$ & 84.88 & 84.77 & 56.00 & 71.94 & 71.86 & 84.00 & 87.80 & 87.78 \\   
& TENT & 40.00${\ast}$ & 75.20 & 75.03 & 42.00${\ast}$ & 76.64 & 76.47 & 56.00${\ast}$ & 88.29 & 88.13 & 72.00${\ast}$ & 86.46 & 86.39 & 64.00 & 75.20 & 75.14 & 84.00 & 89.33 & 89.31 \\
& EATA & 42.00${\ast}$ & 71.92 & 71.77 & 32.00${\ast}$ & 73.93 & 73.72 & 48.00${\ast}$ & 87.44 & 87.24 & 62.00${\ast}$ & 84.85 & 84.74 & 56.00 & 71.94 & 71.86 & 84.00 & 87.81 & 87.79 \\
& SAR & 42.00${\ast}$ & 74.74 & 74.58 & 44.00${\ast}$ & 76.60 & 76.44 & 48.00${\ast}$ & 87.44 & 87.24 & 62.00${\ast}$ & 84.90 & 84.79 & 66.00 & 74.80 & 74.76 & 84.00 & 87.83 & 87.81 \\
& SHOT & 50.00${\ast}$ & 72.35 & 72.24 & 34.00${\ast}$ & 74.57 & 74.37 & 50.00${\ast}$ & 87.58 & 87.39 & 72.00${\ast}$ & 84.53 & 84.47 & 60.00 & 73.54 & 73.47 & 84.00 & 88.81 & 88.79 \\ 
& TEA & 50.00${\ast}$ & 78.30 & 78.16 & 50.00 & 77.89 & 77.75 & 56.00${\ast}$ & 88.86 & 88.70 & 72.00${\ast}$ & 87.15 & 87.07 & 68.00 & 76.69 & 76.65 & 86.00 & 89.39 & 89.37 \\
& MEMO & 64.00 & 67.78 & 67.76  & \textbf{64.00} & 71.05 & 71.01 & \textbf{72.00} & 67.05 & 67.07 & 76.00 & 83.66 & 83.62 & 68.00 & 67.78 & 67.78 & 80.00 & 81.71 & 81.70 \\
\cmidrule(lr){2-20}
& \cellcolor{gray!20}MITA & \textbf{68.00} & \textbf{78.38} & \textbf{78.33} & 54.00 & \textbf{78.29} & \textbf{78.17} & 66.00 & \textbf{89.20} & \textbf{89.08} & \textbf{78.00} & \textbf{87.68} & \textbf{87.63} & \textbf{76.00} & \textbf{79.16} & \textbf{79.14} & \textbf{88.00} & \textbf{89.83} & \textbf{89.82}  \\
\midrule
\multirow{9}*{\parbox[t]{0.7cm}{\centering \rotatebox{90}{Mixed Distributions}}} 
& \cellcolor{pink!20}Source & 45.86 & 28.12 & 36.99 & 41.54 & 34.34 & 37.94 & 65.32 & 53.52 & 59.42 & 73.68 & 74.46 & 74.07 & 27.46  &  28.14 & 27.80 & 52.90 & 58.18 & 55.54 \\
& BN & 57.36 & 70.40 & 63.88 & 64.94 & 62.86 & 63.90 & 80.30 & 86.76 & 83.53 & 71.14${\ast}$  & 84.22 & 77.68 & 61.64  &  70.84 & 66.24 & 85.80 & 86.28 & 86.04  \\   
& TENT & 59.96 & 72.38 & 66.17 & 70.82 & 65.54 & 68.18 & 82.42 & 87.44 & 84.93 & 73.46${\ast}$  & 84.58 & 79.02 & 65.66  &  75.18  & 70.42 & 87.20 & 87.88 & 87.54  \\
& EATA & 57.40 & 70.38 & 63.89 & 64.90 & 63.18 & 64.04 & 80.28 & 86.70  & 83.49 & 71.08${\ast}$  & 84.24 & 77.66 & 61.64 & 70.86 & 66.25 & 85.82 & 86.38 & 86.10 \\
& SAR & 59.76 & 72.32 & 66.04 & 70.08 & 66.76 & 68.42 & 80.28 & 86.68 & 83.48 & 72.82${\ast}$  & 84.76 & 78.79 & 64.64 & 74.06 & 69.35 & 85.80 & 86.28 & 86.04 \\
& SHOT & 62.52 & 71.78 & 67.15 & 71.92  & 70.24 & 71.08 & 80.74 & 85.52 & 83.13 & 69.74${\ast}$  & 81.36 & 75.55 & 61.84 & 69.22 & 65.53 & 86.30 & 87.10 & 86.70 \\ 
& TEA & 64.58 & 74.38 & 69.48 & 75.78  & 72.16 & 73.97 & 82.30 & \textbf{87.62} & 84.96 & 73.80 & 85.46 & 79.63 & 66.64 & 76.56 & 71.60 & 87.50 & 88.58 & 88.04 \\
& MEMO & 62.10 & 69.36 & 65.73 & 65.72 & 71.88 & 68.80 & 82.14 & 67.32 & 74.73 & 73.64${\ast}$  & 84.38 & 79.01 & 59.44 & 69.36 & 64.60 & 81.90 & 83.22 & 82.56 \\
\cmidrule(lr){2-20}
& \cellcolor{gray!20}MITA & \textbf{66.04} & \textbf{74.78} & \textbf{70.41} & \textbf{77.94} & \textbf{73.18} & \textbf{75.56} & \textbf{84.20} & 86.76  & \textbf{85.48} & \textbf{75.78}  & \textbf{85.24} & \textbf{80.51} & \textbf{70.92} & \textbf{78.22} & \textbf{74.57} & \textbf{87.74} & \textbf{88.84} &  \textbf{88.29} \\
\bottomrule
\end{tabular}
\end{adjustbox}
\label{tab:corr_mix}
\end{table*}
\begin{table*}[!t]
\renewcommand{\arraystretch}{0.85}
\caption{Comparisons of MITA and baselines on pure test distributions, using CIFAR-10-C and CIFAR-100-C at the most severe level. The backbone uses WRN-28-10 provided by RobustBench. The best results are highlighted in \textbf{boldface}.}
\centering
\begin{adjustbox}{width=\textwidth}
\setlength{\tabcolsep}{1.2mm}
\begin{tabular}{clccccccccccccccc|cc}
\toprule
\multicolumn{2}{c}{\multirow{2.5}{*}{\centering Dataset}}
& \multicolumn{3}{c}{\textcolor{darkred}{Noise}} 
& \multicolumn{4}{c}{\textcolor{darkgreen}{Blur}} 
& \multicolumn{3}{c}{\textcolor{darkorange}{Weather}} 
& \multicolumn{5}{c}{\textcolor{darkblue}{Digital}} 
& \multicolumn{2}{c}{Avg} \\
\cmidrule(lr){3-5}
\cmidrule(lr){6-9}
\cmidrule(lr){10-12}
\cmidrule(lr){13-17}
\cmidrule(lr){18-19}
& 
& Gauss. & Shot & impulse
& Defoc. & Glass  & Motion & Zoom
& Snow & Frost  & Fog
& Bright. & Contra.  & Elast. & Pixe. & Jpeg.
& Acc & mCE \\
\midrule
\multirow{9}{*}{\parbox[t]{0.7cm}{\centering \rotatebox{90}{CIFAR-10-C}}} 
& \cellcolor{pink!20}Source & 27.68 & 34.25 & 27.07 & 53.01 & 45.67 & 65.24 & 57.99 & 74.87 & 58.68 & 73.98 & 90.70 & 53.37 & 73.39 & 41.56 & 69.71 & 56.47 & 100.00 \\
& BN  & 71.93 & 73.88 & 63.74 & 87.19 & 64.72 & 85.83 & 87.89 & 82.73 & 82.61 & 84.75 & 91.61 & 87.35 & 76.25 & 80.33 & 72.70 & 79.56 & 52.66 \\
& TENT & 75.20 & 76.52 & 67.03 & 88.00 & 68.22 & 86.27 & 89.24 & 84.06 & 83.81 & 86.30 & 92.16 & 87.91 & 78.01 & 82.77 & 75.77 & 81.41 & 48.13 \\
& EATA & 72.25 & 73.88 & 63.74 & 87.19 & 64.72 & 85.83 & 87.89 & 82.73 & 82.61 & 84.75 & 91.61 & 87.35 & 76.25 & 80.33 & 72.70 & 79.59 & 52.63 \\
& SAR & 71.95 & 74.14 & 64.11 & 87.39 & 65.20 & 86.00 & 88.06 & 83.08 & 82.66 & 85.07 & 91.90 & 87.20 & 76.69 & 80.41 & 72.79 & 79.77 & 51.94 \\
& SHOT & 75.19 & 76.66 & 67.16 & 88.12 & 68.70 & 86.56 & 89.19 & 84.37 & 84.18 & 86.56 & 92.28 & 88.65 & 78.18 & 82.82 & 76.05 & 81.64 & 47.47 \\
& TEA & 78.33 & 79.87 & 70.94 & 88.89 & 71.31 & 87.87 &  89.77 & 85.56 & 85.29 & 87.61 & 92.37 & 88.98 & 79.32 & 84.90 & 78.99 & 83.33 & 43.69 \\
& MEMO & 68.75 & 71.69 & 59.13 & 82.81 & 62.05 & 82.23 & 83.78 & 80.88 & 79.58 & 84.40 & 90.49 & 66.84 & 73.46 & 79.11 & 69.97 & 75.68 & 62.13 \\
\cmidrule(lr){2-19}
& \cellcolor{gray!20}MITA & \textbf{79.24} & \textbf{80.68} & \textbf{72.18} & \textbf{89.11} & \textbf{71.79} & \textbf{88.87} & \textbf{90.08} & \textbf{85.84} & \textbf{85.72} & \textbf{87.75} & \textbf{92.59} & \textbf{89.01} & \textbf{79.47} & \textbf{85.67} & \textbf{80.11} & \textbf{83.87} & \textbf{42.37} \\
\midrule
\multirow{9}{*}{\parbox[t]{0.7cm}{\centering \rotatebox{90}{CIFAR-100-C}}}
& \cellcolor{pink!20}Source & 9.87 & 11.58 & 4.15 & 35.57 & 20.56 & 46.92 & 44.20 & 51.13 & 39.07 & 45.07 & 71.42 & 27.43 & 51.01 & 30.19 & 42.82 & 35.39 & 100.00 \\
& BN & 46.53 & 48.62 & 37.15 & 70.94 & 47.36 & 69.04 & 71.25 & 63.00 & 62.96 & 66.08 & 75.89 & 71.31 & 58.79 & 64.56 & 47.46 & 60.06 & 63.54\\
& TENT & 53.44 & 54.12 & 45.53 & 71.69 & 51.17 & 71.54 & 71.63 & 64.88 & 65.30 & 68.41 & 75.14 & 73.59 & 59.25 & 66.81 & 53.93 & 63.09 & 59.42\\
& EATA & 48.76 & 51.60 & 39.47 & 69.29 & 47.49 & 68.13 & 70.68 & 62.94 & 62.53 & 65.14 & 74.48 & 71.61 & 57.53 & 64.24 & 49.67 & 60.24 & 63.75\\
& SAR & 51.34 & 54.08 & 44.62 & 72.24 & 50.10 & 71.06 & 72.43 & 64.96 & 65.35 & 68.40 & 76.23 & 73.95 & 60.04 & 67.26 & 52.29 & 62.95 & 59.37\\
& SHOT & 54.55 & 55.20 & 46.57 & 71.84 & 52.71 & 71.33 & 71.88 & 65.65 & 65.73 & 68.73 & 75.36 & 74.10 & 60.15 & 67.24 & 54.90 & 63.73 & 58.47 \\
& TEA & 54.29 & 56.55 & 48.59 & 72.96 & 53.78 & 72.63 & 74.20 &  67.78 & 67.14 & 69.98 & 76.74 & 75.71 & 62.18 & 68.65 & 55.32 &  65.10 & 56.07 \\
& MEMO & 39.52 & 40.21 & 30.02 & 57.18 & 35.17 & 55.85 & 59.25 & 53.43 & 51.58 & 56.28 & 64.84 & 27.01 & 46.75 & 55.00 & 38.73 & 47.39 & 84.68 \\
\cmidrule(lr){2-19}
& \cellcolor{gray!20}MITA & \textbf{55.10} & \textbf{56.77} & \textbf{49.56} & \textbf{73.12} & \textbf{53.96} & \textbf{72.66} & \textbf{74.25} & \textbf{67.83} & \textbf{67.16} & \textbf{70.22} & \textbf{76.82} & \textbf{75.72} & \textbf{62.37} & \textbf{68.90} & \textbf{56.49} & \textbf{65.40} & \textbf{55.64} \\
\bottomrule
\end{tabular}
\end{adjustbox}
\label{tab:corr_pure}
\end{table*}
\begin{table}[!t]
\renewcommand{\arraystretch}{0.8}
\caption{Ablation Studies of MITA}
\centering
\begin{adjustbox}{width=\linewidth}
\setlength{\tabcolsep}{1.2mm}
\begin{tabular}{l c c c c ccc ccc}
\toprule
\multirow{2}{*}{\parbox[t]{1.8cm}{\centering Component \\ \& Variant}} 
& \multicolumn{2}{c}{Pure Distribution}
& \multicolumn{3}{c}{Outlier} 
& \multicolumn{3}{c}{Mixed Distribution} \\
\cmidrule(lr){2-3} 
\cmidrule(lr){4-6} 
\cmidrule(lr){7-9} 
& Impul. & Gauss.
& Impul. & Gauss. & All 
& Impul. & Gauss. & All  \\
\midrule
Source & 27.07 & 27.68 & 38.00 & 27.66 & 27.72 & 27.46 & 28.14  & 27.80 \\
$\mathrm{MITA_{w/o\;D}}$ & 70.94 & 78.33 & 68.00 & 76.69 & 76.65 & 66.64 & 76.56 & 71.60\\
$\mathrm{MITA_{w/o\;M}}$ & 27.06 & 27.66 & 38.00 & 27.65 & 27.71 & 27.46 & 28.12 & 27.79\\
$\mathrm{MITA_{Same}}$ & 71.24 & 78.89 & 70.00 & 78.31 & 78.27 & 68.98 & 78.14 & 73.56\\
\cmidrule(lr){1-9}
\cellcolor{gray!20}$\mathrm{MITA}$ & \textbf{72.18} & \textbf{79.24} & \textbf{76.00} & \textbf{79.16} & \textbf{79.14}  & \textbf{70.92} & \textbf{78.22}  & \textbf{74.57} \\
\bottomrule
\end{tabular}
\end{adjustbox}
\label{tab:ablation}
\end{table}

\subsection{Experimental Setup}

\paragraph{Datasets and Metrics} 
We use corruption benchmark including CIFAR-10-C and CIFAR-100-C~\cite{krizhevsky2009learning,hendrycks2018benchmarking}, which incorporate 15 types of corruption at five severity levels. 
We use Accuracy and mean Corruption Error (mCE) ~\cite{hendrycks2018benchmarking} as evaluation metrics, applying them to the most severe level of corruption.
\vspace{-2mm}
\paragraph{Baselines} 
We evaluate MITA against seven leading TTA methods: 
BN~\cite{schneider2020improving}, 
TENT~\cite{wang2021tent},  EATA~\cite{niu2022efficient}, SAR~\cite{niu2023towards},
SHOT~\cite{liang2020we}, 
TEA~\cite{yuan2023tea} and
MEMO~\cite{zhang2022memo}.
Notably, ``Source'' denotes the original model without any adaptation.
\vspace{-2mm}
\paragraph{Implementation} 
To ensure a fair comparison, we maintain consistency in model weights following the RobustBench protocol~\cite{croce2021robustbench}, as it provides pre-trained weights for the WRN-28-10~\cite{zagoruyko2016wide} on CIFAR-10.
When RobustBench weights are unavailable, we train the models following~\cite{zagoruyko2016wide}. 
We replicate baselines with their original hyper-parameters if provided. 
More details and setups are provided in~\cref{app:setting}.

\subsection{Adaptation Results}
\label{4.2}

In this section, we evaluate the generalizability of MITA by comparing it with state-of-the-art methods across three distinct scenarios:
(1) An outlier scenario in which the test data predominantly consist of one distribution, with the other distribution appearing as an outlier.
(2) A mixed test distribution scenario where data from two distributions are evenly mixed.
(3) A pure test distribution scenario where the test data consist entirely of a single distribution

Specifically, we create the outlier and mixed distribution scenarios by combining samples from two pure distributions in varying proportions. A highly imbalanced mixing ratio defines an outlier scenario. As the mixing ratio approaches an even balance, the scenario transitions from being an outlier to a mixed distribution scenario. 
In the table below, due to space limitations, we present results for only two mixing ratios, representing outlier and mixture scenarios, respectively. 
For more comprehensive experimental results, please refer to~\cref{app:exp_mix}.

\paragraph{Outlier}

As reported in \cref{tab:corr_mix}, we construct outlier scenarios by mixing samples from two distributions A and B with imbalanced ratios 0.005. For example, a batch may contain 1 outlier from Distribution A and 199 samples from Distribution B. 
For batch-level baseline methods, nearly all perform significantly worse in adapting to these outliers (Dist. A) than even the source model without any adaptation.
For the instance-level method MEMO, while it does not negatively impact the performance on outliers (Dist. A), it struggles with the dominant distribution (Dist. B).
In contrast, our MITA achieves notable improvements in the outlier scenario for both the outliers and dominance.

\paragraph{Mixed Test Distribution}

As reported in \cref{tab:corr_mix}, we construct mixed distributions scenarios by mixing samples from two distributions A and B with balanced ratios 0.5.
MITA significantly outperforms all baselines on both distributions.

\paragraph{Pure Test Distribution}
As reported in~\cref{tab:corr_pure}, we conducted experiments on two datasets against eight baselines for pure test distribution scenario.
MITA markedly surpasses all baselines for all distributions. 
Specifically, MITA outperforms the best-performing baseline by up to 1.74\% for the impulse noise distribution in the CIFAR-10-C dataset.

\paragraph{Effectiveness and Efficiency Analysis}

\begin{figure}[!t]
    \centering
    \includegraphics[width=\linewidth]{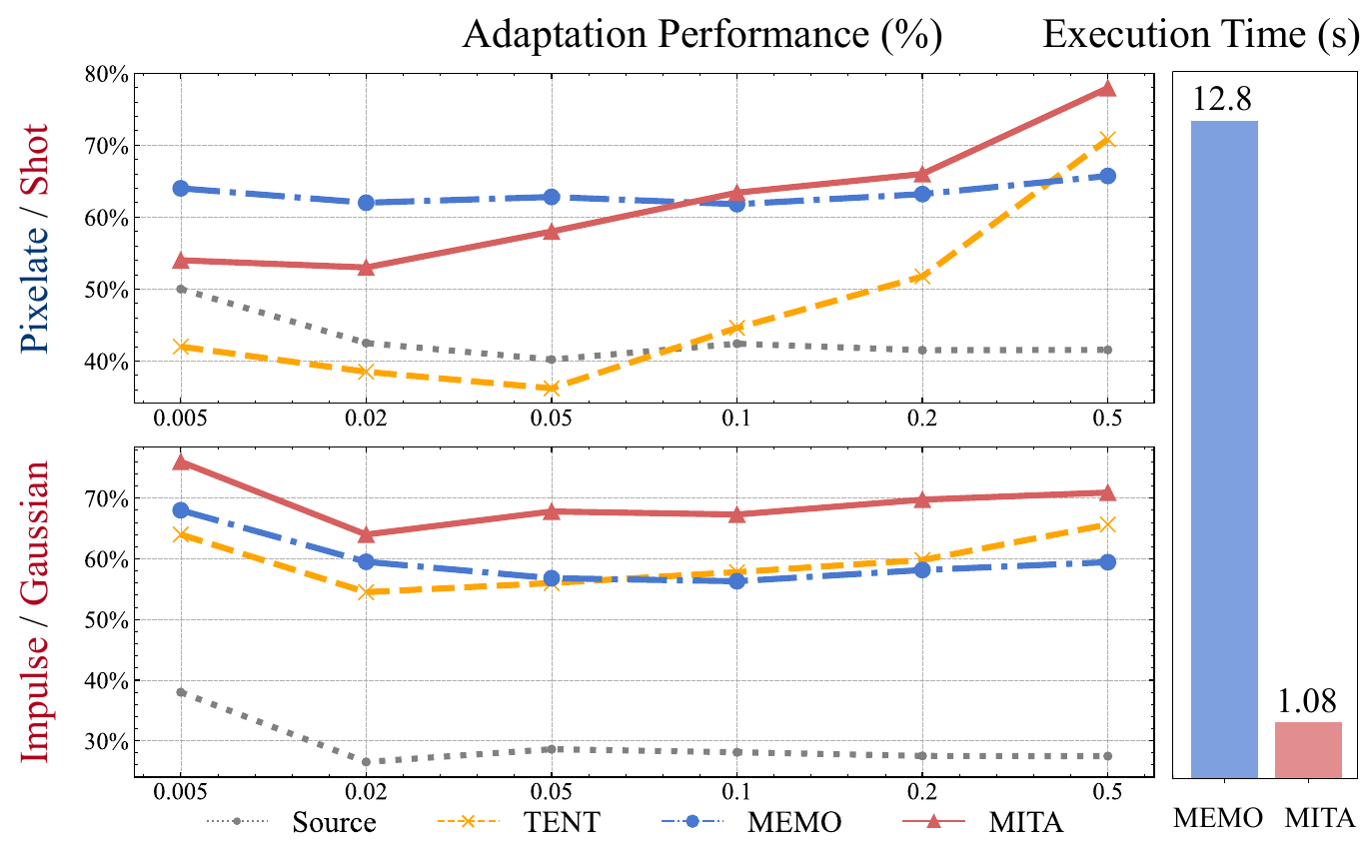}
    \vspace{-5mm}
    \caption{Performance trend on outliers with increasing outlier proportions, ranging from 0.005 to 0.5.}
    \label{fig:trend}
\end{figure}

This section provides further analysis of effectiveness and efficiency across varying distribution ratios as shown in~\cref{fig:trend}.
For significantly different mixed distributions from distinct categories, 
as the ratio increases, batch-level performance gradually improves, yet it does not surpass our method. 
Instance-level performance of MEMO shows no signs of improvement, leading to it initially outperforming but eventually being overtaken by our method.
MEMO's performance is constrained by its inability to utilize knowledge from other samples, especially as the size of similarly distributed data grows, and it also requires significantly more computational time compared to ours.
When the two distributions are from the same category, they may contribute beneficial features to each other, yet both batch and instance-level performances fall short of those achieved by our method.

In summary, Different from batch-level and instance-level methods, MITA bridges the gap between batch-adapted models and single instances, effectively overcoming the batch dependency issue while retaining the ability to utilize knowledge from other samples.

\vspace{-2mm}
\subsection{Ablation Studies}
\label{4.3}

This subsection evaluates the effects of the components or variants of MITA, including:
(1) MITA without both model adaptation and data adaptation, which is the original source model.
(2) MITA without data adaptation, denoted as $\mathrm{MITA_{w/o\;D}}$.
(3) MITA without model adaptation, denoted as $\mathrm{MITA_{w/o\;M}}$.
(4) MITA's data adaptation and model adaptation use the same model, denoted as $\mathrm{MITA_{same}}$.

As shown in \cref{tab:ablation}, our ablation studies confirm three key findings.
Firstly, $\mathrm{MITA_{w/o\;D}}$ outperforms the source model, while $\mathrm{MITA_{w/o\;M}}$ achieves no improvement, underscoring the necessity of model adaptation as a foundation for the whole adaptation.
Secondly, with data adaptation implemented, both $\mathrm{MITA_{same}}$ and MITA surpass the performance of $\mathrm{MITA_{w/o\;D}}$, affirming the effectiveness of data adaptation.
Finally, MITA outperforms $\mathrm{MITA_{Same}}$, underscoring the benefits of prolonged training for an energy-adapted model in enhancing data adaptation.

\subsection{Visualization}

\label{4.4}

In this section, we visualize three aspects:
(1) The model's distribution perceiving ability arising from model adaptation.
(2) The modifications made to the data arising from data adaptation.
(3) Cases where the baseline methods errs, while corrected by MITA under outlier scenario.

\begin{figure}[!t]
    \centering
    \includegraphics[width=0.9\linewidth]{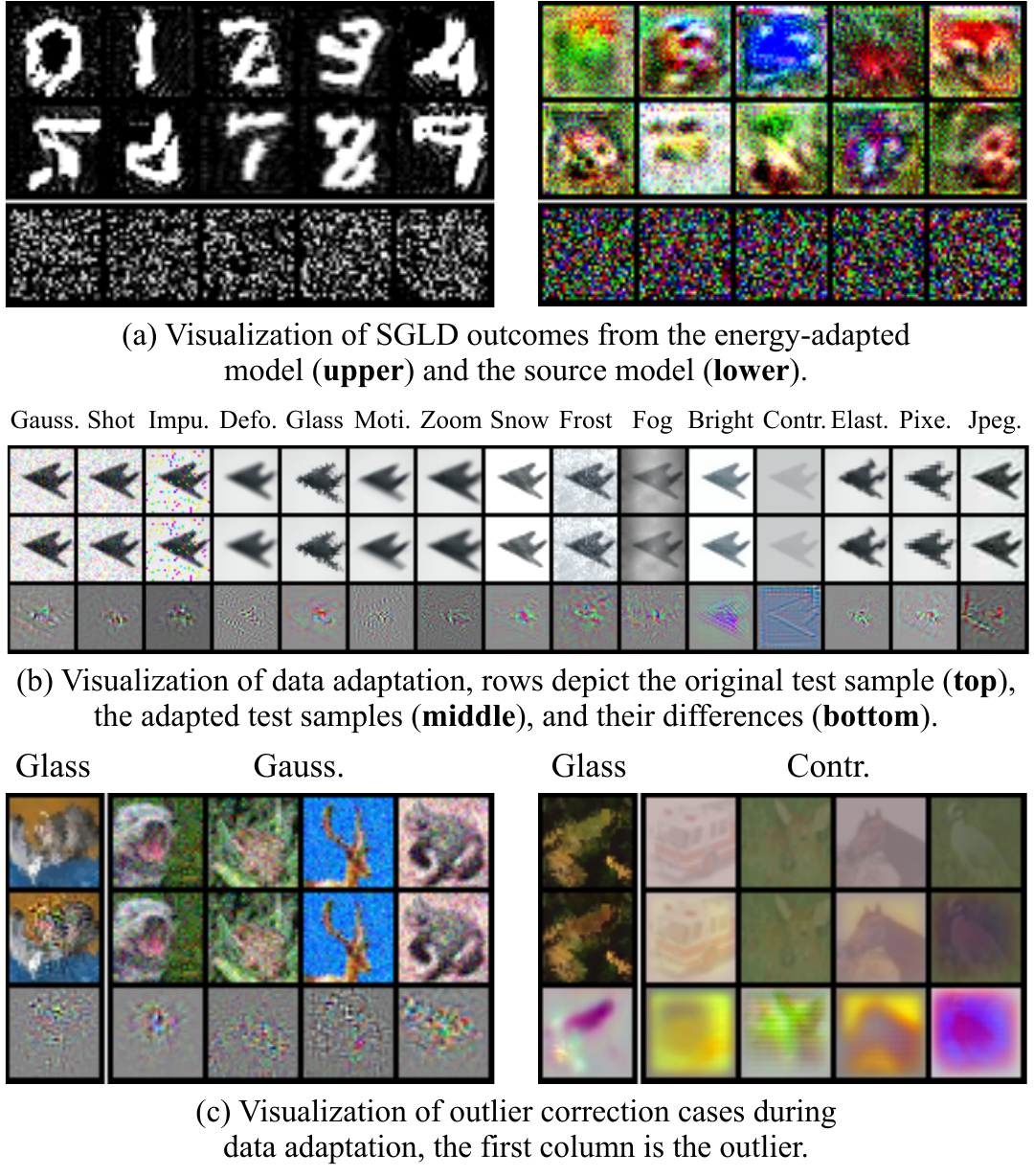}
    \caption{Visualization for model adaptation (a), data adaptation (b) and correction cases during data adaptation (c).}
    \label{fig:vis}
    \vspace{-3mm}
\end{figure}

\paragraph{Model Adaptation}
We employ SGLD to generate samples that visualize the model's embedded distribution \sout{ability} gained from model adaptation. 
As shown in~\cref{fig:vis}~(a), samples drawn from the adapted model can reflect discernible semantic characteristics of the test data.
In contrast, the samples derived from the source model are essentially meaningless noise.
This demonstrates that energy-based model adaptation endows the model with the ability to perceive test distributions and generate relevant data.

\paragraph{Data Adaptation}
We perform MITA on pure test distributions to visualize data adaptation, with results in~\cref{fig:vis}~(b), Each column shows one corruption, and rows depict the original, adapted samples, and their differences.
As can be seen, adapted samples retain the original's semantics, with modifications from adaptation mainly around the focal object. 
Notably, modifications in ``contrast'', ``glass'', and ``brightness'' can exhibit discernible image characteristics.

\paragraph{Correction Cases in Outlier}
We perform MITA under outlier scenario to visualize the  correction cases of data adaptation in~\cref{fig:vis}(c).
As seen, in the left example, the outlier is given noise to match other samples. 
In the right example, the focal object of the outlier is darkened, blending it with the background like the rest with lower contrast.
This demonstrates that the data adaptation process reduces variation within the batch by bringing outliers closer to the main pattern embedded in the model's distribution.

\vspace{-3mm}

\section{Conclusion}
In this work, we introduce MITA (short for Meet-In-The-Middle based Test-Time Adaptation), to bridge the gap between model and data, thus improving model generalizability to distribution shifts. 
Particularly, the proposed MITA involves mutual adaptation for both model and data aspects, which can not only enable the model to capture the overall distributional statistics but also aligns the model with individual instance features that have been neglected by existing TTA methods.
Technically, MITA constructs an energy-based model from the source model and aligns the model's embedded distribution with the test data through unsupervised generative modeling. 
Unlike previous methods that solely align the model towards data,
MITA first constructs a mutual adaptation to make the model and data approach each other. Extensive experiments significantly demonstrate the advantages of MITA and open a new promising paradigm.

\bibliography{main}
\bibliographystyle{icml2024}

\newpage
\appendix
\onecolumn
\section{Appendix Summary}
The appendix contains the following sections:
\begin{itemize}
    \item [(1)] Additional Experiments and Analyses~(\cref{app:exp})
    \begin{itemize}
        \item Extended Settings for Outlier and Mixed Distribution~(\cref{app:exp_mix})
        \item Extended Visualization Results~(\cref{app:exp_vis})
        \item Computing Complexity Analysis~(\cref{app:exp_complexity})
    \end{itemize}
    \item [(2)] Related Works~(\cref{app:relate})
    \item [(3)] Detailed Settings~(\cref{app:setting})
    \begin{itemize}
        \item Datasets~(\cref{app:datasets})
        \item Evaluation Metrics~(\cref{app:metrics})
        \item Hyper-parameters~(\cref{app:hyper})
        \item Computing Resources~(\cref{app:resources})
    \end{itemize}
    \item [(4)] Limitations and Future Explorations~(\cref{app:limitation})
\end{itemize}

\section{Additional Experiments}
\label{app:exp}

\subsection{Extended Settings for Outlier and Mixed Distribution Adaptation}
\label{app:exp_mix}

This section provides more extensive experiments of MITA in scenarios involving outliers and mixed distributions. 
The two scenarios are constructed through the mixing of two distributions with different ratios. 
Scenarios with imbalanced ratios, such as 0.005 and 0.02 for the minority distributions, can be considered as outliers.
As the mixing ratio gradually becomes balanced, such as 0.2 and 0.5, the scenario turns into a mixed distribution.
The results, presented in~\cref{tab:app_corr_mix_1,tab:app_corr_mix_2,tab:app_corr_mix_3,tab:app_corr_mix_4,tab:app_corr_mix_5,tab:app_corr_mix_6} reveal a common trend: while batch-level adaptation methods struggle with the challenges posed by outliers and mixed distributions, MITA demonstrates effective solutions to these issues.

(1) In scenarios with outliers, when the two mixed distributions exhibit significant differences, i.e., originating from different categories as shown in~\cref{tab:app_corr_mix_1,tab:app_corr_mix_2,tab:app_corr_mix_3,tab:app_corr_mix_4}, baseline methods can degrade the performance of the source model. 
In contrast, MITA avoids this pitfall and instead demonstrates improvement.
When the difference between the two mixed distributions is minor, which is the case in~\cref{tab:app_corr_mix_5,tab:app_corr_mix_6}, where both distributions belong to the same category, baseline methods show some improvement. 
However, the enhancement with MITA far exceeds that of all baselines.
These results illustrate that, in cases involving outliers, MITA effectively aligns the model and data, bridging the gap between batch-adapted models and single instances. 
This approach not only addresses the issue of outliers being overshadowed by dominant patterns but also significantly improves the model's performance in terms of dominance.

(2) In scenarios with mixed distributions, MITA significantly outperforms all baseline models on both distributions.
These results further illustrate that, in cases of mixed distributions where the baseline TTA methods may adapt the model to a state that is intermediate between the two distributions, MITA bridges the gap and enables samples from both distributions to better align with the model, thereby enhancing its generalization capabilities.

\begin{table*}[p]
\renewcommand{\arraystretch}{1}
\caption{
Comparisons of MITA and batch-level baselines on outliers and mixtures. The datasets use corruptions of \textbf{Glass} and \textbf{Gaussian} from CIFAR-10-C with significant disparities across categories (\textcolor{darkgreen}{Blur} vs. \textcolor{darkred}{Noise}).
The best adaptation results are highlighted in \textbf{boldface}.}
\centering
\begin{adjustbox}{width=\textwidth}
\setlength{\tabcolsep}{0.5mm}
\begin{tabular}{lcccccccccccccccccc}
\toprule
\multirow{2}{*}{\parbox[t]{2.4cm}{\small \centering Glass / Gaussian \\ (\textcolor{darkgreen}{Blur} / \textcolor{darkred}{Noise})}}
& \multicolumn{3}{c}{0.005} 
& \multicolumn{3}{c}{0.02} 
& \multicolumn{3}{c}{0.05} 
& \multicolumn{3}{c}{0.1} 
& \multicolumn{3}{c}{0.2} 
& \multicolumn{3}{c}{0.5} \\
\cmidrule(lr){2-4}
\cmidrule(lr){5-7}
\cmidrule(lr){8-10}
\cmidrule(lr){11-13}
\cmidrule(lr){14-16}
\cmidrule(lr){17-19}
& Glass & Gauss.  & All
& Glass & Gauss.  & All
& Glass & Gauss.  & All
& Glass & Gauss.  & All
& Glass & Gauss.  & All
& Glass & Gauss.  & All \\
\midrule
\cellcolor{pink!20}Source & 52.00 & 27.67 & 27.79 & 46.50 & 27.71 & 28.09 & 43.40 & 27.62 & 28.41 & 46.10 & 27.67 & 29.51 & 46.70 & 27.58 & 31.40 & 45.86 & 28.12 & 36.99\\
BN & 42.00 & 71.92 & 71.77 & 47.00 & 72.13 & 71.63 & 43.20 & 72.10 & 70.66 & 46.80 & 72.19 & 69.65 & 49.85 & 72.65 & 68.09 & 57.36 & 70.40 & 63.88 \\ 
TENT & 40.00 & 75.20 & 75.03 & 48.50 & 75.15 & 74.62 & 48.20 & 75.48 & 74.12 & 48.40 & 75.22 & 72.54 & 51.95 & 75.21 & 70.56 & 59.96 & 72.38 & 66.17 \\
EATA & 42.00 & 71.92 & 71.77 & 47.50 & 72.15 & 71.66 & 43.00 & 72.13 & 70.67 & 46.80 & 72.16 & 69.62 & 49.90 & 72.66 & 68.11 & 57.40 & 70.38 & 63.89\\
SAR & 42.00 & 74.74 & 74.58 & 48.50 & 74.88 & 74.35 & 47.40 & 75.01 & 73.63 & 49.90 & 74.82 & 72.33 & 52.75 & 74.86 & 70.44 & 59.76 & 72.32 & 66.04 \\
SHOT & 50.00 & 72.35 & 72.24 & 48.50 & 73.61 & 73.11 & 47.80 & 72.66 & 71.42 & 48.00 & 73.84 & 71.26 & 52.80 & 74.56 & 70.21 & 62.52 & 71.78 & 67.15 \\ 
\cellcolor{gray!20}MITA & \textbf{68.00} & \textbf{78.38} & \textbf{78.33} & \textbf{57.50} & \textbf{78.84} & \textbf{78.41} & \textbf{55.00} & \textbf{78.42} & \textbf{77.25} & \textbf{56.50} & \textbf{78.70} & \textbf{76.48} & \textbf{62.05} & \textbf{78.63} & \textbf{74.81} & \textbf{66.04} & \textbf{74.78} & \textbf{70.41} \\
\bottomrule
\end{tabular}
\end{adjustbox}
\label{tab:app_corr_mix_1}
\end{table*}

\begin{table*}[p]
\renewcommand{\arraystretch}{1}
\caption{
Comparisons of MITA and batch-level baselines on outliers and mixtures. The datasets use corruptions of \textbf{Pixelate} and \textbf{Shot} from CIFAR-10-C with significant disparities across categories (\textcolor{darkblue}{Digit} vs. \textcolor{darkred}{Noise}).
The best adaptation results are highlighted in \textbf{boldface}.}
\centering
\begin{adjustbox}{width=\textwidth}
\setlength{\tabcolsep}{0.5mm}
\begin{tabular}{lcccccccccccccccccc}
\toprule
\multirow{2}{*}{\parbox[t]{2.4cm}{\small \centering Pixelate / Shot \\ (\textcolor{darkblue}{Digit} / \textcolor{darkred}{Noise})}}
& \multicolumn{3}{c}{0.005} 
& \multicolumn{3}{c}{0.02} 
& \multicolumn{3}{c}{0.05} 
& \multicolumn{3}{c}{0.1} 
& \multicolumn{3}{c}{0.2} 
& \multicolumn{3}{c}{0.5} \\
\cmidrule(lr){2-4}
\cmidrule(lr){5-7}
\cmidrule(lr){8-10}
\cmidrule(lr){11-13}
\cmidrule(lr){14-16}
\cmidrule(lr){17-19}
& Pixelate & Shot.  & All
& Pixelate & Shot.  & All
& Pixelate & Shot.  & All
& Pixelate & Shot.  & All
& Pixelate & Shot.  & All
& Pixelate & Shot.  & All \\
\midrule
\cellcolor{pink!20}Source & 50.00 & 34.26 & 34.34 & 42.50 & 34.31 & 34.48 & 40.20 & 34.23 & 34.53 & 42.40 & 34.37 & 35.17 & 41.50 & 34.34 & 35.77 & 41.54 & 34.34 & 37.94 \\
BN & 34.00 & 73.82 & 73.62 & 38.00 & 74.01 & 73.29 & 40.20 & 74.13 & 72.43 & 43.00 & 74.13 & 71.02 & 47.65 & 72.91 & 67.86 & 64.94 & 62.86 & 63.90\\
TENT & 42.00 & 76.64 & 76.47 & 38.50 & 76.78 & 76.01 & 36.20 & 77.20 & 75.15 & 44.60 & 77.02 & 73.78 & 51.75 & 76.00 & 71.15 & 70.82 & 65.54 & 68.18 \\
EATA & 32.00 & 73.93 & 73.72 & 38.50 & 74.16 & 73.45 & 36.00 & 74.38 & 72.46 & 39.40 & 74.68 & 71.15 & 52.70 & 73.53 & 69.36 & 64.90 & 63.18 & 64.04 \\
SAR & 44.00 & 76.60 & 76.44 & 39.50 & 76.65 & 75.91 & 40.40 & 76.66 & 74.85 & 46.00 & 76.24 & 73.22 & 51.70 & 75.40 & 70.66 & 70.08 & 66.76 & 68.42 \\
SHOT & 34.00 & 74.57 & 74.37 & 39.00 & 74.88 & 74.16 & 40.60 & 75.03 & 73.31 & 44.90 & 74.90 & 71.90 & 51.75 & 74.25 & 69.75 & 71.92 & 70.24 & 71.08\\ 
\cellcolor{gray!20}MITA & \textbf{54.00} & \textbf{78.29} & \textbf{78.17} & \textbf{53.00} & \textbf{79.75} & \textbf{79.22} & \textbf{58.00} & \textbf{80.01} & \textbf{78.91} & \textbf{63.40} & \textbf{79.99} & \textbf{78.33} & \textbf{66.00} & \textbf{80.05} & \textbf{77.24} & \textbf{77.94} & \textbf{73.18} & \textbf{75.56}  \\
\bottomrule
\end{tabular}
\end{adjustbox}
\label{tab:app_corr_mix_2}
\end{table*}

\begin{table*}[p]
\renewcommand{\arraystretch}{1}
\caption{
Comparisons of MITA and batch-level baselines on outliers and mixtures. The datasets use corruptions of \textbf{Motion} and \textbf{Contrast} from CIFAR-10-C with significant disparities across categories (\textcolor{darkgreen}{Blur} vs. \textcolor{darkblue}{Digit}).
The best adaptation results are highlighted in \textbf{boldface}.}
\centering
\begin{adjustbox}{width=\textwidth}
\setlength{\tabcolsep}{0.4mm}
\begin{tabular}{lcccccccccccccccccc}
\toprule
\multirow{2}{*}{\parbox[t]{2.4cm}{\small \centering Motion / Contrast \\ (\textcolor{darkgreen}{Blur} / \textcolor{darkblue}{Digit})}}
& \multicolumn{3}{c}{0.005} 
& \multicolumn{3}{c}{0.02} 
& \multicolumn{3}{c}{0.05} 
& \multicolumn{3}{c}{0.1} 
& \multicolumn{3}{c}{0.2} 
& \multicolumn{3}{c}{0.5} \\
\cmidrule(lr){2-4}
\cmidrule(lr){5-7}
\cmidrule(lr){8-10}
\cmidrule(lr){11-13}
\cmidrule(lr){14-16}
\cmidrule(lr){17-19}
& Motion & Contra.  & All
& Motion & Contra.  & All
& Motion & Contra.  & All
& Motion & Contra.  & All
& Motion & Contra.  & All
& Motion & Contra.  & All \\
\midrule
\cellcolor{pink!20}Source & 64.00 & 53.35 & 53.41 & 63.00 & 53.40 & 53.59 & 63.20 & 53.39 & 53.88 & 63.80 & 53.38 & 54.42 & 64.60 & 53.35 & 55.60 & 65.32 & 53.52 & 59.42 \\
BN & 48.00 & 87.41 & 87.22 & 60.50 & 87.40 & 86.86 & 60.40 & 87.78 & 86.41 & 64.10 & 87.87 & 85.49 & 71.35 & 88.28 & 84.89 & 80.30 & 86.76 & 83.53 \\   
TENT &  56.00 & 88.29 & 88.13 & 62.00 & 87.91 & 87.39 & 59.40 & 88.64 & 87.18 & 70.80 & 88.80 & 87.00 & 75.35 & 89.12 & 86.37 & 82.42 & \textbf{87.44} & 84.93 \\
EATA & 48.00 & 87.44 & 87.24 & 60.50 & 87.43 & 86.89 & 60.40 & 87.80 & 86.43 & 64.20 & 87.88 & 85.51 & 71.45 & 88.26 & 84.90 & 80.28 & 86.70 & 83.49 \\
SAR & 48.00 & 87.44 & 87.24 & 60.50 & 87.40 & 86.86 & 60.40 & 87.75 & 86.38 & 64.40 & 87.83 & 85.49 & 71.45 & 88.28 & 84.91 & 80.28 & 86.68 & 83.48 \\
SHOT & 50.00 & 87.58 & 87.39 & 56.00 & 87.69 & 87.06 & 58.60 & 88.27 & 86.79 & 67.50 & 88.17 & 86.10 & 74.40 & 87.85 & 85.16 & 80.74 & 85.52 & 83.13 \\ 
\cellcolor{gray!20}MITA &  \textbf{66.00} & \textbf{89.20} & \textbf{89.08} & \textbf{65.50} & \textbf{88.64} & \textbf{88.18} & \textbf{70.00} & \textbf{89.00} & \textbf{88.05} & \textbf{74.20} & \textbf{88.98} & \textbf{87.51} & \textbf{78.40} & \textbf{89.16} & \textbf{87.01} & \textbf{84.20} & 86.76 & \textbf{85.48} \\
\bottomrule
\end{tabular}
\end{adjustbox}
\label{tab:app_corr_mix_3}
\end{table*}

\begin{table*}[p]
\renewcommand{\arraystretch}{1}
\caption{Comparisons of MITA and batch-level baselines on outliers and mixtures. The datasets use corruptions of \textbf{Elastic} and \textbf{Fog} from CIFAR-10-C with significant disparities across categories (\textcolor{darkblue}{Digit} vs. \textcolor{darkorange}{Weather}).
The best adaptation results are highlighted in \textbf{boldface}.}
\centering
\begin{adjustbox}{width=\textwidth}
\setlength{\tabcolsep}{1mm}
\begin{tabular}{lcccccccccccccccccc}
\toprule
\multirow{2}{*}{\parbox[t]{2.4cm}{\small \centering Elastic / Fog \\ (\textcolor{darkblue}{Digit} / \textcolor{darkorange}{Weather})}}
& \multicolumn{3}{c}{0.005} 
& \multicolumn{3}{c}{0.02} 
& \multicolumn{3}{c}{0.05} 
& \multicolumn{3}{c}{0.1} 
& \multicolumn{3}{c}{0.2} 
& \multicolumn{3}{c}{0.5} \\
\cmidrule(lr){2-4}
\cmidrule(lr){5-7}
\cmidrule(lr){8-10}
\cmidrule(lr){11-13}
\cmidrule(lr){14-16}
\cmidrule(lr){17-19}
& Elastic & Fog  & All
& Elastic & Fog  & All
& Elastic & Fog  & All
& Elastic & Fog  & All
& Elastic & Fog  & All
& Elastic & Fog  & All \\
\midrule
\cellcolor{pink!20}Source & 76.00 & 73.96 & 73.98 & 72.00 & 73.99 & 73.95 & 70.40 & 74.01 & 73.83 & 74.70 & 73.94 & 74.02 & 74.05 & 73.86 & 73.90 & 73.68 & 74.46 & 74.07 \\
BN & 62.00 & 84.88 & 84.77 & 63.00 & 85.06 & 84.62 & 60.40 & 84.98 & 83.75 & 65.20 & 84.99 & 83.01 & 67.55 & 85.34 & 81.78 & 71.14 & 84.22 & 77.68 \\   
TENT & 72.00 & 86.46 & 86.39 & 64.50 & 86.41 & 85.97 & 62.80 & 86.18 & 85.01 & 67.40 & 86.23 & 84.35 & 70.00 & 86.38 & 83.11 & 73.46 & 84.58 & 79.02 \\
EATA & 62.00 & 84.85 & 84.74 & 63.00 & 85.12 & 84.68 & 60.60 & 84.97 & 83.75 & 65.20 & 84.99 & 83.01 & 67.45 & 85.38 & 81.79 & 71.08 & 84.24 & 77.66 \\
SAR & 62.00 & 84.90 & 84.79 & 63.00 & 85.09 & 84.65 & 60.40 & 85.00 & 83.77 & 65.20 & 85.02 & 83.04 & 67.75 & 85.56 & 82.00 & 72.82 & 84.76 & 78.79 \\
SHOT & 72.00 & 84.53 & 84.47 & 64.00 & 84.48 & 84.07 & 61.00 & 85.50 & 84.28 & 67.20 & 85.51 & 83.68 & 65.75 & 82.98 & 79.53 & 69.74 & 81.36 & 75.55 \\ 
\cellcolor{gray!20}MITA & \textbf{78.00} & \textbf{87.68} & \textbf{87.63} & \textbf{70.00} & \textbf{87.28} & \textbf{86.93} & \textbf{69.40} & \textbf{87.32} & \textbf{86.42} & \textbf{71.40} & \textbf{87.92} & \textbf{86.27} & \textbf{73.15} & \textbf{86.59} & \textbf{83.90} & \textbf{75.78} & \textbf{85.24} & \textbf{80.51} \\
\bottomrule
\end{tabular}
\end{adjustbox}
\label{tab:app_corr_mix_4}
\end{table*}

\begin{table*}[!t]
\renewcommand{\arraystretch}{1}
\caption{
Comparisons of MITA and batch-level baselines on outliers and mixtures. The datasets use corruptions of \textbf{Impulse} and \textbf{Guassian} from CIFAR-10-C with greater similarity within the same \textcolor{darkred}{Noise} category.
The best adaptation results are highlighted in \textbf{boldface}.}
\centering
\begin{adjustbox}{width=\textwidth}
\setlength{\tabcolsep}{0.2mm}
\begin{tabular}{lcccccccccccccccccc}
\toprule
\multirow{2}{*}{\parbox[t]{2.4cm}{\small \centering Impulse / Guassian \\ (\textcolor{darkred}{Noise} / \textcolor{darkred}{Noise})}}
& \multicolumn{3}{c}{0.005} 
& \multicolumn{3}{c}{0.02} 
& \multicolumn{3}{c}{0.05} 
& \multicolumn{3}{c}{0.1} 
& \multicolumn{3}{c}{0.2} 
& \multicolumn{3}{c}{0.5} \\
\cmidrule(lr){2-4}
\cmidrule(lr){5-7}
\cmidrule(lr){8-10}
\cmidrule(lr){11-13}
\cmidrule(lr){14-16}
\cmidrule(lr){17-19}
& Impul. & Gauss.  & All
& Impul. & Gauss.  & All
& Impul. & Gauss.  & All
& Impul. & Gauss.  & All
& Impul. & Gauss.  & All
& Impul. & Gauss.  & All \\
\midrule
\cellcolor{pink!20}Source & 38.00  & 27.66  &  27.72  & 26.50  &  27.71  & 27.69  & 28.60 &  27.62 &  27.67  & 28.10  &  27.67  &  27.71  & 27.50 & 27.57 &   27.56  & 27.46  &  28.14 & 27.80 \\
BN &  56.00  & 71.94 &  71.86  & 50.00 & 71.92 &  71.48 & 52.60 & 71.92  &  70.95  & 53.30  &  72.19  & 70.30 & 55.60 & 72.38 &  69.02 & 61.64  &  70.84 & 66.24   \\   
TENT &   64.00 &  75.20  &  75.14  & 54.50 &  75.22 & 74.81 & 56.00 & 75.43  &  74.46  & 57.80  &  75.13 &   73.40 & 59.80 & 75.04  &   71.99 & 65.66  &  75.18  &    70.42   \\
EATA & 56.00 & 71.94 & 71.86 & 50.00 & 71.96 & 71.53 & 52.60 & 71.94 & 70.98 & 53.30 & 72.22 & 70.33 & 55.65 & 72.36 & 69.02 & 61.64 & 70.86 & 66.25 \\
SAR &  66.00 & 74.80 & 74.76 & 52.50 & 74.89 & 74.44 & 55.80 & 74.78 & 73.84 & 57.10 & 74.82 & 73.05 & 59.45 & 74.91 & 71.82 & 64.64 & 74.06 & 69.35 \\
SHOT & 60.00 & 73.54 & 73.47 & 56.50 & 72.91 & 72.58 & 56.40 & 73.59 & 72.73 & 57.00 & 72.21 & 70.69 & 59.70 & 72.48 & 69.92 & 61.84 & 69.22 & 65.53 \\ 
\cellcolor{gray!20}MITA & \textbf{76.00} & \textbf{79.16} & \textbf{79.14} & \textbf{64.00} & \textbf{78.72} & \textbf{78.43} & \textbf{67.80} & \textbf{78.80} & \textbf{78.25} & \textbf{67.30} & \textbf{78.89} & \textbf{77.73} & \textbf{69.75} & \textbf{78.93} & \textbf{77.09} & \textbf{70.92} & \textbf{78.22} & \textbf{74.57} \\
\bottomrule
\end{tabular}
\end{adjustbox}
\label{tab:app_corr_mix_5}
\end{table*}

\begin{table*}[!t]
\renewcommand{\arraystretch}{1}
\caption{
Comparisons of MITA and batch-level baselines on outliers and mixtures. The datasets use corruptions of \textbf{Defocus} and \textbf{Zoom} from CIFAR-10-C with greater similarity within the same \textcolor{darkgreen}{Blur} category.
The best adaptation results are highlighted in \textbf{boldface}.}
\centering
\begin{adjustbox}{width=\textwidth}
\setlength{\tabcolsep}{0.5mm}
\begin{tabular}{lcccccccccccccccccc}
\toprule
\multirow{2}{*}{\parbox[t]{2.4cm}{\small \centering Defocus / Zoom \\ (\textcolor{darkgreen}{Blur} / \textcolor{darkgreen}{Blur})}}
& \multicolumn{3}{c}{0.005} 
& \multicolumn{3}{c}{0.02} 
& \multicolumn{3}{c}{0.05} 
& \multicolumn{3}{c}{0.1} 
& \multicolumn{3}{c}{0.2} 
& \multicolumn{3}{c}{0.5} \\
\cmidrule(lr){2-4}
\cmidrule(lr){5-7}
\cmidrule(lr){8-10}
\cmidrule(lr){11-13}
\cmidrule(lr){14-16}
\cmidrule(lr){17-19}
& Defoc. & Zoom  & All
& Defoc. & Zoom  & All
& Defoc. & Zoom  & All
& Defoc. & Zoom  & All
& Defoc. & Zoom  & All
& Defoc. & Zoom  & All \\
\midrule
\cellcolor{pink!20}Source & 50.00 & 57.95 & 57.91 & 54.00 & 57.98 & 57.90 & 51.40 & 58.02 & 57.69 & 51.50 & 57.98 & 57.33 & 51.75 & 58.10 & 56.83 & 52.90 & 58.18 & 55.54 \\
BN & 84.00 & 87.80 & 87.78 & 86.00 & 87.90 & 87.86 & 84.00 & 87.89 & 87.70 & 84.20 & 87.90 & 87.53 & 84.10 & 87.90 & 87.14 & 85.80 & 86.28 & 86.04 \\   
TENT &  84.00 & 89.33 & 89.31 & 86.50 & 89.12 & 89.07 & 85.60 & 89.31 & 89.12 & 84.70 & 89.30 & 88.84 & 85.75 & 88.94 & 88.30 & 87.20 & 87.88 & 87.54 \\
EATA & 84.00 & 87.81 & 87.79 & 86.00 & 87.88 & 87.84 & 84.00 & 87.88 & 87.69 & 84.20 & 87.90 & 87.53 & 84.20 & 87.86 & 87.13 & 85.82 & 86.38 & 86.10 \\
SAR & 84.00 & 87.83 & 87.81 & 86.00 & 87.84 & 87.80 & 84.00 & 87.89 & 87.70 & 84.20 & 87.90 & 87.53 & 84.10 & 87.90 & 87.14 & 85.80 & 86.28 & 86.04 \\
SHOT & 84.00 & 88.81 & 88.79 & 88.00 & 88.80 & 88.78 & 85.40 & 88.77 & 88.60 & 84.30 & 88.74 & 88.30 & 85.10 & 88.58 & 87.88 & 86.30 & 87.10 & 86.70 \\ 
\cellcolor{gray!20}MITA & \textbf{88.00} & \textbf{89.83} & \textbf{89.82} & \textbf{90.50} & \textbf{89.76} & \textbf{89.77} & \textbf{87.00} & \textbf{89.82} & \textbf{89.68} & \textbf{87.90} & \textbf{89.96} & \textbf{89.75} & \textbf{87.05} & \textbf{89.31} & \textbf{88.86} & \textbf{87.74} & \textbf{88.84} & \textbf{88.29} \\
\bottomrule
\end{tabular}
\end{adjustbox}
\label{tab:app_corr_mix_6}
\end{table*}

\subsection{Extended Visualization Results}
\label{app:exp_vis}

\begin{figure}[!t]
    \centering
    \includegraphics[width=\linewidth]{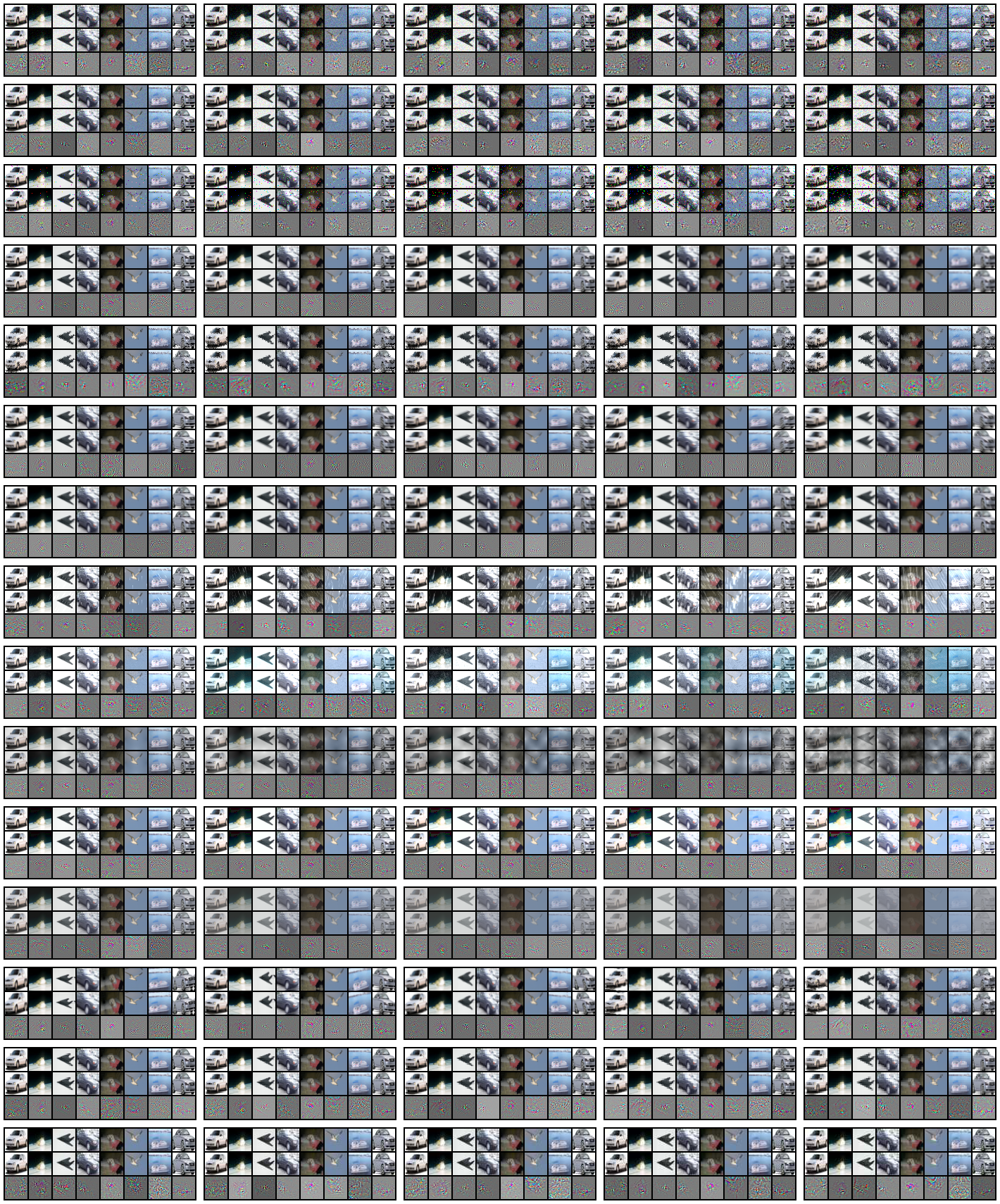}
    \caption{Visualization of data adaptation sample modifications across various types of corruption. Each of the 15 rows corresponds to a different corruption type, arranged from top to bottom as follows: gaussian, shot, impulse, defocus, glass, motion, zoom, snow, frost, fog, brightness, contrast, elastic, pixelate, and jpeg. The five columns represent increasing severity levels from 1 to 5, moving from left to right.}
    \label{fig:app_vis_bn}
\end{figure}

This section presents additional visual results of data adaptation sample modifications. As shown in~\cref{fig:app_vis_bn},
two key observations about the data adaptation process stand out. 
First, the process maintains the semantic content of the images, ensuring that the essence and interpretability are unaffected despite the presence of noise or distortions. 
Second, the adaptations are not uniformly distributed but are instead focused on critical regions, such as the subjects and their outlines. 
This targeted modification underscores the model's ability to enhance important features within an image, boosting the generalizability to various distribution shifts.

\subsection{Computing Complexity Analysis}
\label{app:exp_complexity}

This experiment quantitatively compares the computational complexity of batch-level, instance-level, and our proposed method. The experimental settings are as follows: 
the dataset consists of the first 200 samples from the CIFAR-10-C with Gaussian noise at level five. 
We employed WideResNet-28-10 as the backbone architecture, and all tests were conducted on the same NVIDIA A800 GPU. 
The metrics used in this comparison include the counts of parameters, FLOPs, and the execution time for processing 200 samples.

As shown in ~\cref{tab:app_flops}, although it is higher than the batch-level methods, it is significantly better than the instance-level method. 
Notably, even in cases where the FLOPs are similar (e.g., MITA with sgld-step of 5 and MEMO with batch-size of 16), our method's support for batch parallel operations results in a performance improvement of approximately 91.56\% in execution time compared to MEMO.
In summary, given the significant generalization performance improvements of MITA over both batch-level and instance-level methods, its computational complexity is deemed acceptable.

\begin{table}[!h]
\renewcommand{\arraystretch}{0.6}
\caption{\footnotesize Comparison of Parameter Counts, FLOPs and Execution Time for 200 samples}
\centering
\begin{adjustbox}{width=0.5\linewidth}
\begin{tabular}{clccccc}
\toprule
& Method & Params (M) & FLOPs (G) & Time (s) \\
\midrule
& Source & 36.48 & 5.25 & 0.03 \\
\midrule
\multirow{3}{*}{\parbox[t]{1cm}{\centering Batch}}
& TENT & 36.48 & 5.25 & 0.07 \\
& LAME & 36.48 & 10.50 & 0.03 \\
& TEA & 36.48  & 21.01  & 0.24 \\
\midrule
\multirow{3}{*}{\parbox[t]{1cm}{\centering Instance}}
& MEMO (bs=4) & 36.48 & 26.26 & \textcolor{red}{8.80}\\
& MEMO (bs=8) & 36.48 & 47.27 & \textcolor{red}{9.60} \\
& MEMO (bs=16) & 36.48 & 89.29 & \textcolor{red}{12.80} \\
\midrule
\multirow{2}{*}{\parbox[t]{1cm}{\centering Ours}}
& MITA (sgld-step=1) & 72.95 & 42.02 & \textbf{0.44}\\
& MITA (sgld-step=5) & 72.95 & 84.04 & \textbf{1.08} \\
\bottomrule
\end{tabular}
\end{adjustbox}
\label{tab:app_flops}
\end{table}

\section{Related Works}
\label{app:relate}

\paragraph{Test Time Adaptation}
Test Time Adaptation (TTA)~\cite{liang2023comprehensive} is a paradigm aiming to enhance a model's generalizability on specific test data through unsupervised fine-tuning with these data. 
Note that the model is originally trained on a distinct training dataset, and during this adaptation phase, neither the original training data nor the training process is available, which is particularly beneficial for large, open-source models where training details are often proprietary or resource-constrained~\cite{touvron2023llama,xiao2024Cal,xiao2024how}.
Approaches like TTT~\cite{pmlr-v119-sun20b} adapt models through self-supervised proxy task during testing but require the training of the same proxy task during training procedure. 
DDA~\cite{Gao_2023_CVPR, xiao2023energybased} explores adapting the test data, yet faces limitations due to model structure and training constraints.

Recent research~\cite{wang2021tent} highlights a scenario where the training process and training data is entirely agnostic, leading to five main categories of approaches:
For normalization-based methods,
BN~\cite{schneider2020improving} adapts the BatchNorm~\cite{ioffe2015batch} statistics with test data.
DUA~\cite{mirza2022norm} uses a tiny fraction of test data and its augmentation for BatchNorm statistics adaptation.
For entropy-based methods, 
TENT~\cite{wang2021tent} fine-tunes BatchNorm layers using entropy minimization during the test phase.
EATA~\cite{niu2022efficient} employs a Fisher regularizer to limit excessive model parameter changes.
SAR~\cite{niu2023towards} removes high-gradient samples and promotes flat minimum weights.
For consistency based methods, 
MEMO~\cite{zhang2022memo} enforces invariance across augmentations for each given test sample.
For pseudo-labeling-based, 
PL~\cite{lee2013pseudo} fine-tunes parameters using confident pseudo labels.
SHOT~\cite{liang2020we} combines entropy minimization methods with pseudo labeling.
There are also unclassified methods: T3A~\cite{iwasawa2021test} modifies the last layer by the pseudo-prototype representations of each class. LAME~\cite{Boudiaf_2022_CVPR} adapts the output via a Laplacian regularization. 

\paragraph{Energy-Based Models}
Energy-Based Models (EBMs) are a type of non-normalized probabilistic models. Unlike most other probabilistic models~\cite{hou2022augmentation,hou2022conditional}, 
EBMs do not necessitate the normalizing constant to be tractable~\cite{lecun2006tutorial,song2021train} and do not require an explicit neural network for sample generation, implying the generation process is implicit~\cite{du2019implicit}.
These lead to increased flexibility in parameterization and allow for modeling a wider range of probability distributions.
Due to their flexibility, EBMs can construct hybrid models with both discriminative and generative capabilities, integrating the generative competencies into discriminative models without sacrificing their discriminative capabilities~\cite{Grathwohl2020Your,du2019implicit,han2019divergence}. Among these, JEM~\cite{Grathwohl2020Your} is particularly representative, reinterpreting classifiers as an EBM and achieving impressive results in both classification and generation.

\section{Detailed Settings}
\label{app:setting}

\subsection{Datasets}
\label{app:datasets}

We perform experiments on four datasets across two tasks. Image corruption task include CIFAR-10-C, CIFAR-100-C, and ImageNet-C datasets. 
Domain generalization task include DomainNet datasets.

\paragraph{Dataset of Corrupted Distributions}
CIFAR-10-C, CIFAR-100-C and ImageNet-C~\cite{hendrycks2018benchmarking} are variants of the original CIFAR-10, CIFAR-100 and Tiny-ImageNet datasets that have been artificially corrupted into 15 types of corruptions at five levels of severity, resulting in 75 corrupted versions of the original test set images. 
The corruptions include 15 main corruptions: Gaussian noise, shot noise, impulse noise, defocus blur, glass blur, motion blur, zoom blur, snow, frost, fog, brightness, contrast, elastic, pixelation, and JPEG.
All these corruptions are simulations of shifted distributions that models might encounter in real-world situations.

\paragraph{DomainNet}
DomainNet~\cite{neyshabur2020being} is a dataset of common objects in six different domain. All domains include 345 classes of objects. 
The domains include clipart, real, sketch, infograph, painting and quickdraw.

\begin{table}[!h]
\centering
\caption{Summary of Corruption Datasets}
\begin{adjustbox}{width=0.5\linewidth}
\setlength{\tabcolsep}{2mm}
\begin{tabular}{lcccc}
\toprule
\textbf{Dataset} & \textbf{\#Sample} & \textbf{\#Corr.} & \textbf{\#Severity} & \textbf{\#Class.} \\
\midrule
CIFAR-10-C & 10,000$\times$15$\times$5 & 15 &5 & 10 \\
CIFAR-100-C  & 10,000$\times$15$\times$5 & 15 &5 & 100  \\
ImageNet-1000-C & 50,000$\times$15$\times$5 & 15 &5 & 1000  \\
\bottomrule
\end{tabular}
\end{adjustbox}
\label{Tab:Dataset_Corr}
\end{table}

\subsection{Evaluation Metrics}
\label{app:metrics}
Following previous works, we employ Average Accuracy and Mean Corruption Error (mCE)~\cite{hendrycks2018benchmarking,yuan2023pde,yuan2022towards} as evaluation metrics for corruption datasets. 
\paragraph{Average Accuracy} Average Acc is the accuracy averaged over all severity levels and corruptions. Consider there are a total of $C$ corruptions, each with $S$ severities. For a model $f$, let $\mathcal{E}_{s,c}(f)$ denote the top-1 error rate on the corruption $c$ with severity level $s$ averaged over the whole test set,
\begin{equation}
    \mathrm{AverAcc}_f=1-\frac{1}{C \cdot S} \sum_{c=1}^{C} \sum_{s=1}^S \mathcal{E}_{s, c}(f).
\end{equation}
\vspace{-3mm}
\paragraph{Mean Corruption Error} mCE is a metric used to measure the performance improvement of model $f$ compared to a baseline model $f_0$. We use the model without adaptation as the baseline model,

\begin{equation}
\mathrm{mCE}_f=\frac{1}{C} \sum_{c=1}^C \frac{\sum_{s=1}^S \mathcal{E}_{c, s}(f)}{\sum_{s=1}^S \mathcal{E}_{c, s}\left(f_0\right)} 
\end{equation}

\subsection{Hyper-parameters}
\label{app:hyper}
Our hyperparameter selection consists of empirical initialization and fine-tuning via optuna~\cite{akiba2019optuna} search. 
For the empirical initialization, we align with the configurations used in Tent~\cite{wang2021tent} with respect to common hyperparameters, such as those pertaining to optimizers. 
As for MITA-specific hyperparameters, we adhere to the parameter choices from TEA~\cite{yuan2023tea} and subsequently employ optuna for fine-grained search.

\subsection{Computing resources}
\label{app:resources}
All our experiments are performed on RedHat server (4.8.5-39) with Intel(R) Xeon(R) Gold 5218 CPU $@$ 2.30GHz4, $4 \times$ NVIDIA Tesla V100 SXM2 (32GB) and $3 \times$ NVIDIA Tesla A800 SXM4 (80GB).

\section{Limitation and Future Works}
\label{app:limitation}

Our study has identified key aspects for improvement and future research, which are outlined below:

(1) Computational Complexity and Stability: The use of SGLD is time-consuming and sometimes unstable. 
However, ongoing research in energy-based models is addressing these challenges through various methods. These include gradient clipping~\cite{yang2021jem++}, diffusion processes~\cite{luo2023training}, the introduction of an additional gradient term~\cite{pmlr-v139-du21b}, and sampling based on ordinary differential equations (ODE)~\cite{nie2021controllable}. 
One of our future directions is to enhance the efficiency of MITA by incorporating these advanced sampling techniques.

(2) Transferability and Discriminability Trade-Off:
The trade-off between transferability and discriminability is a common issue in TTA research~\cite{kundu2022balancing, Gao_2023_CVPR}. 
In the context of MITA, transferability refers to the model's generative capacity. 
However, overemphasizing this capacity could significantly impair the model's discriminability, which necessitates the use of two separate energy models, one for model adaptation and the other for data adaptation.
Therefore, another important direction for our future work is to explore how to enhance the model's transferability while maintaining or even improving its discriminative power.

We acknowledge that the identified limitations may present challenges.
Nevertheless, we remain confident that our study represents a pioneering step to bridge the gap between model and data, and we believe that these limitations could be addressed in future efforts.

\end{document}